%% file: main.tex
\documentclass[preprint,12pt,authoryear]{elsarticle}

\usepackage{amssymb}
\usepackage{amsmath}

\usepackage{lineno}

\newcommand{\algrule}[1][.2pt]{\par\vskip.5\baselineskip\hrule height #1\par\vskip.5\baselineskip}
\newcommand{\thinalgrule}[1][.2pt]{\par\vskip.3\baselineskip\hrule height #1\par\vskip.3\baselineskip}
\newcommand{\company}{OVS}

\usepackage{color,soul} 
\usepackage{algorithm, algpseudocode} 
\usepackage[table]{xcolor} 
\usepackage{subcaption} 
\usepackage{booktabs} 
\usepackage{listings} 
\usepackage{csvsimple} 
\usepackage{chato-notes}
\usepackage[noadjust]{marginnote}
\usepackage{hyperref} 
\usepackage{orcidlink} 
\usepackage{pgfplots} 
\pgfplotsset{compat=1.18} 
\usepackage{pgfplotstable} 
\usepackage{filecontents} 
\usepackage{tikz} 
\usepackage{listofitems} 
\usetikzlibrary{shapes.geometric} 
\usetikzlibrary{patterns} 
\usepgfplotslibrary{fillbetween} 

\pgfdeclareplotmark{mystar}{
    \node[star,star point ratio=2.25,minimum size=6pt,
          inner sep=0pt,draw=black,solid,fill=black] {};
}

\newcommand{\review}[1]{{\color{black} {#1}}}

\journal{}

\begin{document}

\begin{frontmatter}



\title{Training-Free, Identity-Preserving Image Editing for Fashion Pose Alignment and Normalization}


\author[poliba,wideverse]{Potito Aghilar\corref{cor}}\ead{potito.aghilar@poliba.it, potito.aghilar@wideverse.com}
\author[poliba]{Vito Walter Anelli}\ead{vitowalter.anelli@poliba.it}
\author[wideverse]{Michelantonio Trizio}\ead{michelantonio.trizio@poliba.it, michelantonio.trizio@wideverse.com}
\author[poliba]{Eugenio Di Sciascio}\ead{eugenio.disciascio@poliba.it}
\author[poliba]{Tommaso Di Noia}\ead{tommaso.dinoia@poliba.it}

\cortext[cor]{Corresponding author}


\affiliation[poliba]{organization={Polytechnic University of Bari},
            addressline={Via Edoardo Orabona, 4}, 
            city={Bari},
            postcode={70125}, 
            state={Apulia},
            country={Italy}}
\affiliation[wideverse]{organization={Wideverse},
            addressline={Via Edoardo Orabona, 4}, 
            city={Bari},
            postcode={70125}, 
            state={Apulia},
            country={Italy}}


\begin{abstract}
\input{0_abstract}
\end{abstract}




\end{frontmatter}












\input{1_introduction}
\input{2_method}

\input{3_experiments}
\input{4_limitations}

\input{5_conclusion}
\section*{Acknowledgements}


The authors acknowledge partial support of the following projects: OVS: Fashion Retail Reloaded, Lutech Digitale 4.0, Secure Safe Apulia, BIO-D, Meditech, EPANSA (FAIR) - Enhancing Personal Assistants with Neuro-Symbolic AI and Knowledge Graphs, funded by the European Union Next- GenerationEU (Italian National Recovery and Resilience Plan (NRRP) – M4C2, Investment 1.3, D.R. No. 123 of 16/01/2024, PE00000013, CUP: H97G22000210007). We acknowledge the CINECA award under the ISCRA initiative for the availability of high-performance computing resources and support. We acknowledge ISCRA for awarding this project access to the LEONARDO supercomputer, owned by the EuroHPC Joint Undertaking, hosted by CINECA (Italy). 

\bibliographystyle{elsarticle-harv}
\bibliography{main}




\end{document}

%% file: 0_abstract.tex
Diffusion models have recently unlocked new possibilities in editing images of real-world objects. Yet, transforming objects in non-rigid ways, such as modifying poses or applying image-based conditioning, continues to present significant challenges. Retaining the unique identity of objects during these edits is a complex task, and current techniques often fall short of delivering the precision needed for industrial settings, where consistency is non-negotiable. Additionally, adapting diffusion models demands custom training data, which is often unavailable in real-world scenarios. To address these gaps, we present \textsc{FashionRepose}, a novel, training-free pipeline designed to handle non-rigid pose adjustments specifically for the fashion industry. This approach combines pretrained off-the-shelf models to modify the poses of long-sleeve garments while safeguarding their identity and branding characteristics. By adopting a zero-shot methodology, \textsc{FashionRepose} enables near real-time edits, entirely eliminating the requirement for specialized training data. \textsc{FashionRepose} has been deployed for a global fashion firm, \textsf{\textsc{\company}}, handling more than 30{,}000 long-sleeve garments.

%% file: 1_introduction.tex
\section{Introduction}
\label{sec:introduction}

The digital transformation of the fashion industry has placed a growing emphasis on image editing, which has become a cornerstone of e-commerce, marketing, and design. 
Fashion brands today heavily rely on the ability to modify garment poses, alter colors, and visualize new styles for diverse audiences. These capabilities streamline workflows, reduce costs, and provide enhanced customer experiences~\citep{mohammadi-2021}. However, the challenge of achieving non-rigid transformations -- particularly pose adjustments or normalization~\citep{DBLP:journals/tii/ZhaoDHM23} -- remains unresolved. This is especially critical for maintaining brand identity, which relies on consistency in visual representation. The increasing demand for personalized and engaging content has only intensified the pressure on brands to adopt efficient yet precise image editing techniques that can scale with consumer expectations.

Traditional graphic tools like Photoshop~\footnote{Photoshop - \url{https://www.adobe.com/it/products/photoshop}} and Illustrator~\footnote{Illustrator - \url{https://www.adobe.com/it/products/illustrator}} have been the industry standard for decades, enabling precise but highly manual editing workflows~\citep{hume2020fashion, Caruso2002-ku, altenburg2014gimp, hume2020fashion}. While effective, these tools demand significant time and expertise, making scalability a persistent issue. Editors need to iterate over numerous design variants manually, a task that becomes increasingly laborious as product lines expand or marketing campaigns diversify. This bottleneck can hinder a brand's ability to remain agile when faced with rapidly changing trends. In contrast, advancements in Artificial Intelligence (AI) and Machine Learning (ML) have introduced automated methods capable of generating photorealistic virtual samples in minutes. These innovations shorten product design cycles, enhance collaboration between designers and manufacturers, and allow for quicker responses to market trends~\citep{Ramos2023}. The ability to create virtual prototypes on demand has also enabled fashion companies to experiment with innovative concepts without the financial and logistical constraints of traditional photoshoots.

AI-driven tools have been developed to facilitate a variety of tasks, including image editing, search, and recommendation~\citep{DBLP:conf/um/Aghilar25,DBLP:conf/recsys/Palma23,DBLP:conf/recsys/AnelliNLS17,DBLP:conf/um/AnelliBNBTS17}. These tools enable fashion brands to adapt to evolving consumer preferences, seasonal trends, and marketing strategies~\citep{guo-2023}. However, despite these advances, existing AI-based image editing solutions often fall short when it comes to delivering precise, identity-preserving edits. Non-rigid transformations, such as garment pose adjustments, are particularly challenging due to the complexity of maintaining garment identity, texture fidelity, and branding consistency. These shortcomings limit the broader adoption of AI-based tools for professional and industrial applications, like automatic import of entire products catalog in virtual showrooms. 

These limitations become clear when current AI techniques are examined. Generative Adversarial Networks (GANs)~\citep{DBLP:journals/corr/GoodfellowPMXWOCB14}, characterized by a dual-network setup with a generator and a discriminator, are able to produce results that can appear photorealistic. Despite their ability to generate high quality outputs, GANs struggle with maintaining subject identity during non-rigid transformations, such as garment pose modifications. Probabilistic diffusion models (DMs), which generate images through an iterative denoising process, have emerged as a superior alternative, offering greater detail and flexibility~\citep{DBLP:conf/cvpr/RombachBLEO22, DBLP:conf/nips/HoJA20, DBLP:conf/cvpr/RombachBLEO22, DBLP:conf/iclr/PerniasRRPA24, DBLP:conf/icml/EsserKBEMSLLSBP24, DBLP:journals/corr/abs-2308-06721}. Unlike GANs, DMs generate higher-quality results with fewer artifacts, ideal for controlled image editing tasks.

Conditioning techniques developed for DMs allow fine-grained control over generated images by using inputs like text descriptions, sketches, or poses~\citep{DBLP:journals/corr/abs-2308-06721}. These advancements have led to applications such as Stable Diffusion, which supports text-conditioned image generation, and ControlNet, which extends these capabilities to pose-controlled generation~\citep{DBLP:conf/iccv/ZhangRA23}. In fashion applications, these tools enable creative experimentation with garment designs and poses\review{~\citep{DBLP:conf/iccv/DongLSWLZH019}}. However, their reliance on specific conditioning inputs and the need for model fine-tuning can limit their adaptability for diverse and rapidly evolving industrial use cases.

The practical deployment of GANs and diffusion models in real-world fashion applications is further hindered by their reliance on fine-tuning and extensive domain-specific training datasets. Moreover, in many cases, the absence of domain-specific or task-specific datasets makes re-training or fine-tuning impossible. Techniques like Low-Rank Adaptation (LoRA) improve the efficiency of fine-tuning~\citep{DBLP:conf/iclr/HuSWALWWC22}, but they still require significant computational resources and time, which are impractical for fast-paced industries. Inversion methods, such as GAN inversion~\citep{DBLP:journals/pr/PernusFSD25} and DDIM/DPM inversion~\citep{DBLP:conf/iclr/SongME21}, attempt to estimate the latent noise used to generate original images, offering a pathway to precise edits. However, these methods frequently fail to preserve garment identity, particularly when used on Out-Of-Distribution (OOD) images. Additionally, the dependence on textual prompts or other rigid control mechanisms can hinder the expressiveness and adaptability needed for nuanced editing tasks. These constraints highlight the need for a more flexible, efficient, and scalable approach through the definition of a production-ready pipeline~\citep{DBLP:conf/ecsa/AghilarATN23, DBLP:journals/tii/DaiZKCH23}.

To overcome these challenges, we propose \textsc{FashionRepose}, a novel training-free pipeline specifically designed for garment pose alignment in long-sleeve clothing. In contrast to Virtual Try-On systems that simulate a garment on a person or mannequin, our task is to simulate garment sleeve repositioning on the same still-life image while maintaining identity and brand-specific attributes. By leveraging pretrained AI models and computer vision algorithms, \textsc{FashionRepose} achieves consistent pose editing without the need for additional training data, through a zero-shot methodology.  
The training-free emerging approach refers to methods that leverage pre-trained models, heuristics, or search-based strategies to perform tasks without requiring additional training or fine-tuning~\citep{DBLP:journals/tog/TewelKGKWCA24, DBLP:conf/cvpr/XuYWWSS24,DBLP:conf/iclr/Zhang0J0Z024,DBLP:conf/iccv/YuWZGZ23}. These approaches have gained traction in various fields, including machine learning, computer vision~\citep{DBLP:conf/aaai/WeiDHLLLP024, DBLP:conf/cvpr/ZhouSZLST0J22}, and automated decision-making, due to their ability to provide effective solutions while avoiding the computational and data requirements of traditional training-based methods.
The adoption of training-free approaches is driven by several key motivations. These methods eliminate the high computational and financial costs associated with training large models, making AI more sustainable and accessible. They reduce dependence on large labeled datasets, enabling deployment in data-scarce environments. Additionally, training-free techniques enhance adaptability, allowing models to generalize across tasks without retraining. They also offer faster deployment, which is crucial for real-time applications, and provide greater interpretability compared to deep learning's "black-box" nature. Finally, these approaches democratize AI by making powerful solutions viable in low-resource settings, ensuring broader accessibility and practical usability.
\textsc{FashionRepose} preserves garment identity and branding attributes while eliminating the delays and inefficiencies associated with fine-tuning and retraining. Moreover, the pipeline is designed to operate seamlessly across diverse long sleeve garment types, making it both scalable and adaptable to industry needs. The proposed solution integrates cutting-edge advancements in diffusion models into a unified end-to-end pipeline that addresses a specific fashion use case.

Unlike conventional methods that rely heavily on textual prompts or fine-tuned datasets, \textsc{FashionRepose} emphasizes adaptability and speed, making it particularly suited for real-world applications where scalability and consistency are paramount. Its near real-time editing capabilities ensure that fashion brands can meet the demands of dynamic markets without compromising quality or brand identity.


The contributions of our study are summarized as follows:
\begin{enumerate}
    \item A zero-shot pose alignment pipeline tailored for long-sleeve garments, providing efficient and consistent edits suitable for industrial applications.
    \item A computer vision algorithm for identifying and segmenting sleeve and torso regions, enabling precise and consistent edits while preserving garment structure.
    \item A logo-preservation workflow that detects, suppresses, and reintegrates branding elements, such as logos and labels, to maintain brand identity across edited images.
\end{enumerate}

By addressing the limitations of current AI techniques, \textsc{FashionRepose} offers a scalable, precise, and identity-preserving solution for the fashion industry. Its potential applications span e-commerce, marketing, and design prototyping, enabling fashion brands to produce high-quality, real-time image transformations that meet the needs of modern consumers or B2B use cases. To the best of our knowledge, \textsc{FashionRepose} is the first approach proposing a fully integrated, zero-shot pipeline that effectively aligns fashion garments while explicitly preserving critical brand features. Moreover, both the training-free and zero-shot approaches ensure the proposed solution can adapt to future challenges, positioning it as a versatile tool in the ever-evolving landscape of digital fashion innovation. By fostering creativity, enhancing operational efficiency, and maintaining visual consistency, \textsc{FashionRepose} represents a significant leap forward for AI-powered fashion technologies.


%% file: 2_method.tex
\section{Methodology}
\label{sec:method}

\textsc{FashionRepose} is a zero-shot pipeline designed for garment pose alignment, preserving visual fidelity without requiring training. It efficiently manages non-rigid transformations while ensuring garment identity.

\subsection{Task Definition and Motivation}
\label{sec:task_definition}

We formally define the garment pose alignment task as follows. Given an input image \(I_{in}\), representing a garment in a still-life pose, the objective is to generate an output image, \(I_{out}\), with the garment repositioned to a target pose, while preserving identity, texture, and branding features.

Formally, the task is described as

\begin{equation}
I_{out} = \mathcal{T}(I_{in}, \theta_{target}),
\end{equation}

where:
\begin{itemize}
  \item \(\mathcal{T}\) denotes the pose transformation function.
  \item \(\theta_{target} \in [0, 90]^\circ\) represents the target arm-to-torso angle.
\end{itemize}

In the following, the goal is to transform long sleeve garments from a relaxed arm configuration (\textit{Still-Life Pose}) to a novel pose (\textit{Target Pose}), without affecting other image elements:
\begin{itemize}
    \item \textbf{the Still-Life Pose}: Initial position with arms relaxed close to the body, pointing towards the ground.
    \item \textbf{the Target Pose}: Desired configuration with arms raised at a specific arm-to-torso angle, ranging from 0 to 90 degrees. When the target pose is set to 45 degrees, we define the pose as \textit{Normalized Pose}.
\end{itemize}
Traditional pose editing models rely on retraining techniques and annotated datasets, limiting their efficiency for real-time fashion applications that demand real-time, consistent, and high-quality edits. Diffusion models often distort garment details, such as shape and texture, during transformations. Moreover, the lack of diverse clothing poses datasets hinders domain-specific training. Our training-free, zero-shot pipeline resolves these issues by achieving real-time garment pose alignment tailored for practical use in the fashion industry while preserving integrity and addressing occlusions.
\textsc{FashionRepose} is not only a theoretically novel solution but also one that has been practically validated in a demanding industrial setting. Deployed within the production workflow of \textsf{\textsc{\company}}~\footnote{OVS - \url{https://www.ovscorporate.it/en/about-us/ovs-group}}, a global fashion retailer, it has significantly improved the speed and cost-efficiency of product photoshoots at scale, demonstrating its robustness, real-world applicability, and alignment with the operational needs of the fashion industry.

\subsection{Pipeline Architecture}
\label{sec:pipeline_architecture}
\review{The pose alignment pipeline processes a single input still-life image of a long sleeve garment to obtain a reposed image while preserving critical attributes, including color, texture, and branding. This pipeline, designed to promote a zero-shot and training-free approach, operates with high efficiency, achieving results in less than one minute. The architecture is presented in Figure~\ref{fig:pose_norm_pipeline}, while the data flow is summarized in Algorithm~\ref{alg:pose_norm_pipeline}.} The pipeline comprises the following main stages:
\begin{figure*}[t]
    \centering
    \includegraphics[width=\linewidth]{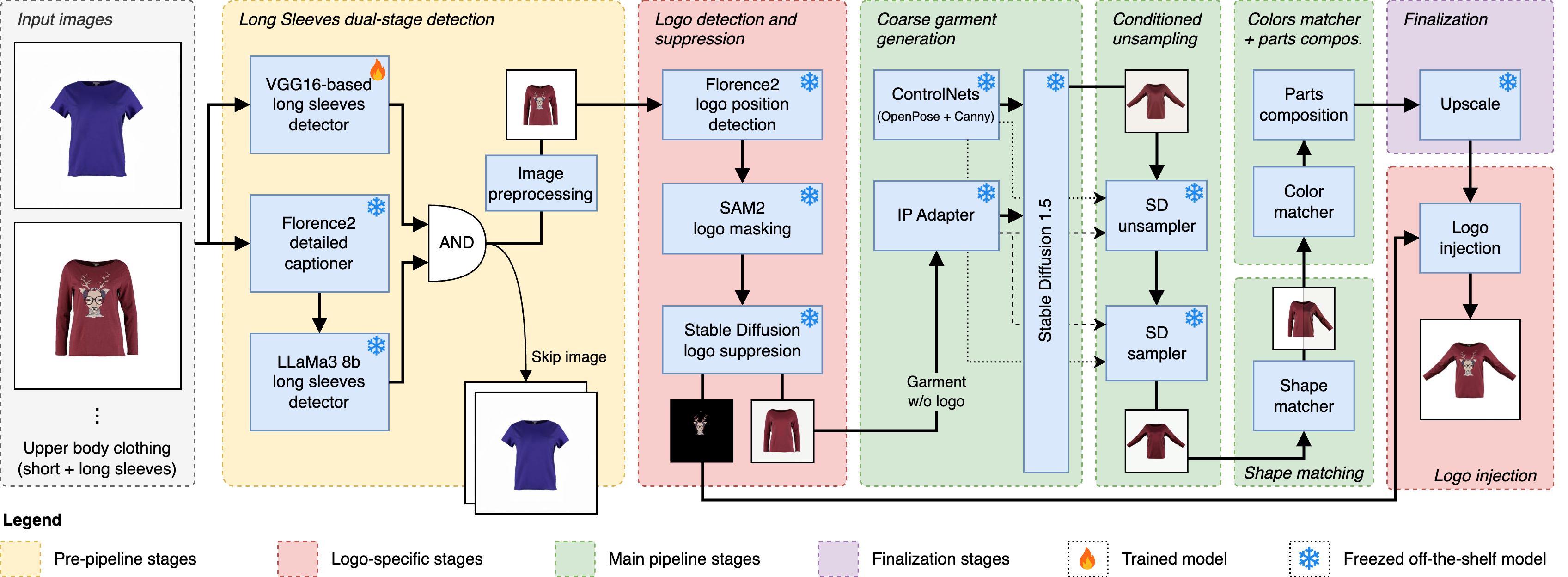}
    \caption{\textbf{Architecture of \textsc{FashionRepose}}. The training-free pipeline is capable of editing, in less than a minute, the pose of long-sleeved garments, maintaining consistency and identity. The workflow is entirely based on pretrained off-the-shelf models and computer vision techniques, making it easy to integrate into existing industry processes.}
    \label{fig:pose_norm_pipeline}
\end{figure*}
\begin{itemize}
    \item \textbf{Long Sleeves Detection}: Detects and filters out all non-long-sleeve garments images. 
    \item \textbf{Logo Detection and Suppression}: Detects and suppress all the logos and labels in the input image.
    \item \textbf{Coarse Generation}: Generates a coarse aligned-pose garment using pretrained diffusion models. 
    \item \textbf{Conditioned Unsampling}: Applies conditioned noise to move the garment within the latent space. 
    \item \textbf{Source-Target Shape Matching}: Performs a mask-guided alignment using the shapes of the garments. 
    \item \textbf{Garment Parts-Composition}: Blends the torso from the original image with the generated sleeves. 
    \item \textbf{Logo Injection}: Restores all the logos and labels to retain brand identity across edits. 
\end{itemize}
Each stage of the pipeline has been carefully designed to address the challenges of achieving accurate, visually consistent edits within minimal processing time. These capabilities are essential in fashion applications, where real-time performance and identity retention are two critical aspects.

\begin{algorithm}[h!]
  \review{
  \footnotesize
  \caption{\review{FashionRepose pseudocode}}
  \label{alg:pose_norm_pipeline}
  \begin{algorithmic}[1]
    \State \textbf{Input:} Still-life garment image $I_{in}$, target pose angle $\theta_{target}$.
    \State \textbf{Output:} Reposed, identity-preserved garment image $I_{out}$.
    \thinalgrule
    \vspace{0.2em}
    \Statex \textbf{\textit{Long Sleeves Detection}}
    \vspace{0.4em}
    \State $F \leftarrow $ identify non-long-sleeve garment from $I_{in}$;
    \If {$F$ is \textbf{False}} 
        \State \textbf{Return}; \Comment{Abort execution}
    \EndIf
    \thinalgrule
    \vspace{0.2em}
    \Statex \textbf{\textit{Logo Detection and Suppression}}
    \vspace{0.4em}
    \State $I_{ls}, I_{logos}, M_{logos} \leftarrow$ extract and suppress logos from $I_{in}$;
    \thinalgrule
    \vspace{0.2em}
    \Statex \textbf{\textit{Coarse Generation}}
    \vspace{0.4em}
    \State $S_{pose}, M_{pose} \leftarrow$ generate target skeleton pose and mask starting from $\theta_{target}$;
    \State $E_{ip}, E_{pose}, E_{canny} \leftarrow$ extract embeddings from $I_{ls}$, $S_{pose}$, $M_{pose}$;
    \State $I_{coarse} \leftarrow$ generate coarse garment from $E_{ip}, E_{pose}, E_{canny}$;
    \thinalgrule
    \vspace{0.2em}
    \Statex \textbf{\textit{Conditioned Unsampling}}
    \vspace{0.4em}
    \State $Z_{ls}, Z_{coarse} \leftarrow$ encode latents from $I_{ls}$ and $I_{coarse}$;
    \State $Z_{blend} \leftarrow$ blend latents $Z_{ls}$ and $Z_{coarse}$ using gradient-based mask $Z_{gradient}$;
    \State $\widetilde{E}_{pose}^{u}, \widetilde{E}_{canny}^{u}, \widetilde{E}_{pose}^{s}, \widetilde{E}_{canny}^{s} \leftarrow$ compute embeddings to guide the diffusion process;
    \State $I_{refined} \leftarrow$ perform the Conditioned Unsampling process;
    \thinalgrule
    \vspace{0.2em}
    \Statex \textbf{\textit{Source-Target Shape Matching}}
    \vspace{0.4em}
    \State $M_{ls}, M_{refined} \leftarrow$ segment masks from $I_{ls}$ and $I_{refined}$;
    \State $\overline{I}_{ls}, \overline{I}_{refined} \leftarrow$ align $I_{ls}$ and $I_{refined}$ using $M_{ls}$, $M_{refined}$ and $M_{pose}$;
    \thinalgrule
    \vspace{0.2em}
    \Statex \textbf{\textit{Garment Parts-Composition}}
    \vspace{0.4em}
    \State $M^{@45}_{pose} \leftarrow$ generate fixed 45$^\circ$ garment composition mask using $\mathcal{P}$;
    \State $I_{composed} \leftarrow$ compose torso from $\overline{I}_{ls}$ and sleeves from $\overline{I}_{refined}$ using $M^{@45}_{pose}$;
    \thinalgrule
    \vspace{0.2em}
    \Statex \textbf{\textit{Logo Injection}}
    \vspace{0.4em}
    \State $I^{\uparrow}_{composed} \leftarrow$ upscale $I_{composed}$ to final target resolution via $\mathcal{U}_{\uparrow}$;
    \State $I_{out} \leftarrow$ restore logos into $I^{\uparrow}_{composed}$ using $I_{logos}$ and $M_{logos}$;
    \State \textbf{Return} $I_{out}$.
  \end{algorithmic}
  }
\end{algorithm}

\input{2_stage1}
\input{2_stage2}
\input{2_stage3}
\input{2_stage4}
\input{2_stage5}
\input{2_stage6}
\input{2_stage7}

%% file: 2_stage1.tex
\subsection{Long Sleeves Detection}
\label{sec:long_sleeves_detection}

In line with recent works in fashion image classification task~\citep{DBLP:journals/eswa/KolisnikHZ21, DBLP:journals/eswa/SeoS19}, the detection of long sleeve garments constitutes a pivotal preprocessing step designed to exclude images that do not feature upper-body garments with long sleeves. This step optimizes the pipeline by ensuring that only relevant images advance to subsequent stages, thereby improving overall efficiency and accuracy. 

We formally define the long sleeves detection task as follows. Given an input still-life image \(I_{in}\), the goal is to determine whether it depicts a long sleeve garment. This is achieved through a two-step filtering process:

\begin{itemize}
    \item A deep learning-based classifier, \(\mathcal{D}_{VGG16}\), fine-tuned on a subset of DressCode~\citep{DBLP:journals/tog/HeYZYLX24}, predicts the presence of long sleeves.
    \item A semantic refiner, \(\mathcal{D}_{semantic}\), validates image captions generated by a vision captioner component, \(\mathcal{F}_{caption}\), given by a combination of Florence2~\citep{DBLP:conf/cvpr/0004WXDHL00Y24} and LLaMA3~\citep{DBLP:journals/corr/abs-2407-21783}.
\end{itemize}

\noindent The two classifiers operate as follows:
\begin{equation}
F_{VGG16} = \mathcal{D}_{VGG16}(I_{in}), \quad F_{VGG16} \in \{0,1\},
\end{equation}
\begin{equation}
F_{semantic} = \mathcal{D}_{semantic}(\mathcal{F}_{caption}(I_{in})), \quad F_{semantic} \in \{0,1\},
\end{equation}
where $F_{VGG16}$ and $F_{semantic}$ are the outcomes of the two classifiers. 
The final decision is made by combining the outputs,
\begin{equation}
F = F_{VGG16} \land F_{semantic},
\end{equation}
where \(\land\) denotes the logical AND operator.

Although not essential, these filtering steps improve the pipeline by removing irrelevant images, allowing for seamless integration into real-world workflows, excluding human intervention. Moreover, the downstream, training-free editing stages align with the goal of efficient, zero-shot garment pose alignment, critical for practical fashion industry applications.


\begin{table}[t]
    \centering
    \footnotesize
    \rowcolors{2}{gray!20}{}
    \setlength{\tabcolsep}{5pt}
    \begin{tabular}{@{}lccc@{}}
        \toprule
        \textbf{Dataset partition} & \textbf{Training (80\%)} & \textbf{Validation (20\%)} & \textbf{Total} \\
        \midrule
        Upper body (long sleeve) & 480 & 120 & 600 \\
        Upper body (short sleeve) & 480 & 120 & 600 \\
        Lower body & - & - & - \\
        Dresses & - & - & - \\
        \midrule
        \textbf{Total} & 960 & 240 & 1200 \\
        \bottomrule
    \end{tabular}
    \caption{\textbf{Dataset Split for Long Sleeve Detection stage. }DressCode~\citep{DBLP:journals/tog/HeYZYLX24} dataset splitting for long sleeve detection stage (see Sec.~\ref{sec:long_sleeves_detection}). We manually sampled from the dataset the most representative images for each class.}
    \label{tab:dataset_split}
\end{table}

\begin{table}[t]
    \centering
    \footnotesize
    \rowcolors{2}{gray!20}{}
    \setlength{\tabcolsep}{2.5pt}
    \begin{filecontents*}{results_fixed.csv}
Model,Optimizer,Accuracy,Precision,Recall,F1 Score
VGG16 *,RMSprop *,\textbf{0.9669},\textbf{1.0000},\underline{0.9322},\textbf{0.9649}
VGG16,AdamW,\underline{0.9545},\textbf{1.0000},0.9068,\underline{0.9511}
ResNet50,RMSprop,0.9504,0.9829,0.9200,0.9504
ResNet50,AdamW,0.9504,0.9829,0.9200,0.9504
Vanilla CNN,RMSprop,0.9380,\underline{0.9912},0.8898,0.9378
Vanilla CNN,AdamW,0.9380,\underline{0.9912},0.8898,0.9378
InceptionV3,RMSprop,0.9380,0.9322,\textbf{0.9402},0.9362
InceptionV3,AdamW,0.9339,0.9316,0.9316,0.9316
\end{filecontents*}
\csvreader[
        tabular=@{}cccccc@{},
        table head=\toprule \textbf{Model} & \textbf{Optimizer} & \textbf{Accuracy ($\uparrow$)} & \textbf{Precision ($\uparrow$)} & \textbf{Recall ($\uparrow$)} & \textbf{F1 ($\uparrow$)}\\\midrule,
        late after line=\\,
        late after last line=\\\midrule\multicolumn{6}{c}{* Best model selected for \textit{Long Sleeves Detection} stage in the pipeline}\\\bottomrule
    ]
    {results_fixed.csv}
    {Model=\model,Optimizer=\optimizer,Accuracy=\accuracy,Precision=\precision,Recall=\recall,F1 Score=\fscore}
    {\model & \optimizer & \accuracy & \precision & \recall & \fscore }
    \caption{\textbf{Metrics for Long Sleeve Detection module.} Quantitative metrics for long sleeve detection module integrated into the pipeline. Best results are reported in \textbf{bold}, while second-best are \underline{underlined}.}
    \label{tab:sleeve_detection_results}
\end{table}

We evaluated multiple state-of-the-art deep learning architectures, including VGG16~\citep{DBLP:journals/corr/SimonyanZ14a}, ResNet50~\citep{resnet50}, and InceptionV3~\citep{inception_v3}, alongside a vanilla CNN model. These models are chosen based on their architectural diversity and proven effectiveness in different classification tasks. VGG16, with its uniform stacking of 3x3 convolutional filters, offers simplicity and strong feature extraction capabilities. ResNet50 leverages residual learning to overcome vanishing gradient issues in deeper networks, while InceptionV3 incorporates multi-scale parallel filters, allowing it to capture features at varying levels of granularity. The CNN serves as a baseline, undergoing full training from randomly initialized weights, providing a benchmark for the effectiveness of pretrained models. Due to the absence of detailed annotations for long-sleeve garments, the DressCode dataset’s upper-body garment category was manually split into two subsets: long-sleeve and non-long-sleeve garments. Each subset contains 600 images, resulting in a balanced dataset of 1200 samples (see Table~\ref{tab:dataset_split}). To ensure representativeness, the images are selected through cherry-picking. The fine-tuning process leverages the pretrained weights of each model, updating only the final layers while keeping earlier layers frozen to retain general feature extraction capabilities. In contrast, the vanilla CNN undergoes full training from scratch, which provides a useful comparison point for evaluating the effectiveness of transfer learning. All the models are trained using the binary cross-entropy with logits loss:
\begin{equation}
\label{eq:binary_cross_entropy_loss_1}
    loss(x, y) = mean(\{l_1,..., l_N\})
\end{equation}
where $loss$ is the per-batch loss, \( N \) is the batch size and the loss for each sample in the batch \( l_n \) is defined as:
\begin{equation}
\label{eq:binary_cross_entropy_loss_2}
    l_n = -w_n [ y_n \cdot log\;\sigma(x_n) + (1 - y_n) \cdot log\;(1 - \sigma(x_n)) ]
\end{equation}
where $x_n$ and $y_n$ are the predicted logits for the n-th sample in the batch and the ground truth label for the n-th sample, respectively. 
$w_n$ is an optional weighting term that adjusts the contribution of the n-th sample to the overall loss. This is often used for handling imbalanced datasets. We proved in our analysis that VGG16 is the best performing model in this task (see Table~\ref{tab:sleeve_detection_results}).

The semantic refiner is a subsequent filtering step we introduced to filter out the images that are not properly detected by the first step. It is composed by a Florence2-based image captioner and a LLaMA3 detector in which we asked to check if the garment description contains or not long sleeves.

%% file: 2_stage2.tex
\subsection{Logo Detection and Suppression}
\label{sec:logo_detection_suppression}

\begin{figure}[t]
    \centering
    \includegraphics[width=\linewidth]{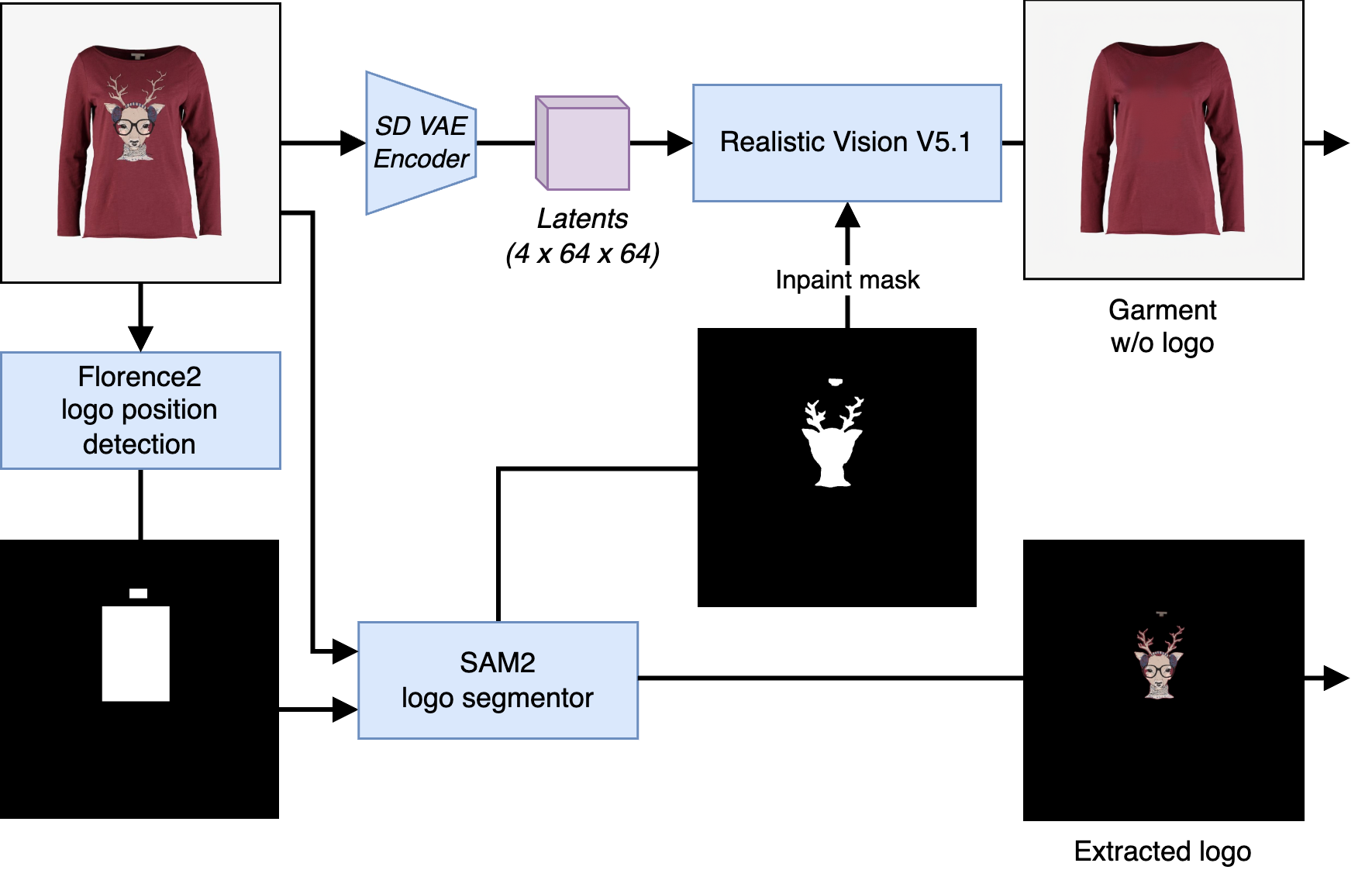}
    \caption{\textbf{Logo Detection and Suppression.} Starting from the original clothing image, Florence2~\citep{DBLP:conf/cvpr/0004WXDHL00Y24} is applied to detect the bounding boxes for all logos and labels. SAM2~\citep{DBLP:journals/corr/abs-2408-00714} then isolates and extracts these components from the image. Finally, a diffusion-based inpainting method fills the gaps within the garment.}
    \label{fig:logo_detection_pipeline}
\end{figure}

To preserve the garment identity, the pipeline incorporates a logo restoration phase, starting with the detection and suppression of logos and labels in the original still-life image prior to editing. Figure~\ref{fig:logo_detection_pipeline} illustrates this process.
We formally define the logo suppression step as follows. Given an input image \(I_{in}\), the task is to remove the logos and obtain a clean version of the image, while preserving the logo information for later reinjection.
\begin{equation}
I_{ls}, I_{logos}, M_{logos} = \mathcal{L}(I_{in}),
\end{equation}
where \(\mathcal{L}\) is the logo suppression function, \(I_{ls}\) is the resulting logo-suppressed image, \(I_{logos}\) is the image containing the extracted logo regions, and \(M_{logos}\) are the corresponding segmentation masks used during suppression.

\paragraph{\textbf{Logo Detection}} For this phase, we employed a two-step approach: Florence2~\citep{DBLP:conf/cvpr/0004WXDHL00Y24} initially localizes all the logos and labels by providing bounding boxes, subsequently SAM2~\citep{DBLP:journals/corr/abs-2408-00714} refines them into precise segmentation masks. This dual-stage approach ensures high accuracy and robustness during logo detection.

\paragraph{\textbf{Logo Suppression}} The suppression phase performs an image inpainting operation utilizing the segmentation masks generated during the detection phase. This operation allows for seamless removal of the logo by reconstructing the underlying garment texture and color within the masked region. An additional image morphology dilation is applied to the mask to enhance the inpainting result and avoid artifacts.

%% file: 2_stage3.tex
\begin{figure}[t]
    \centering
    \includegraphics[width=0.8\linewidth]{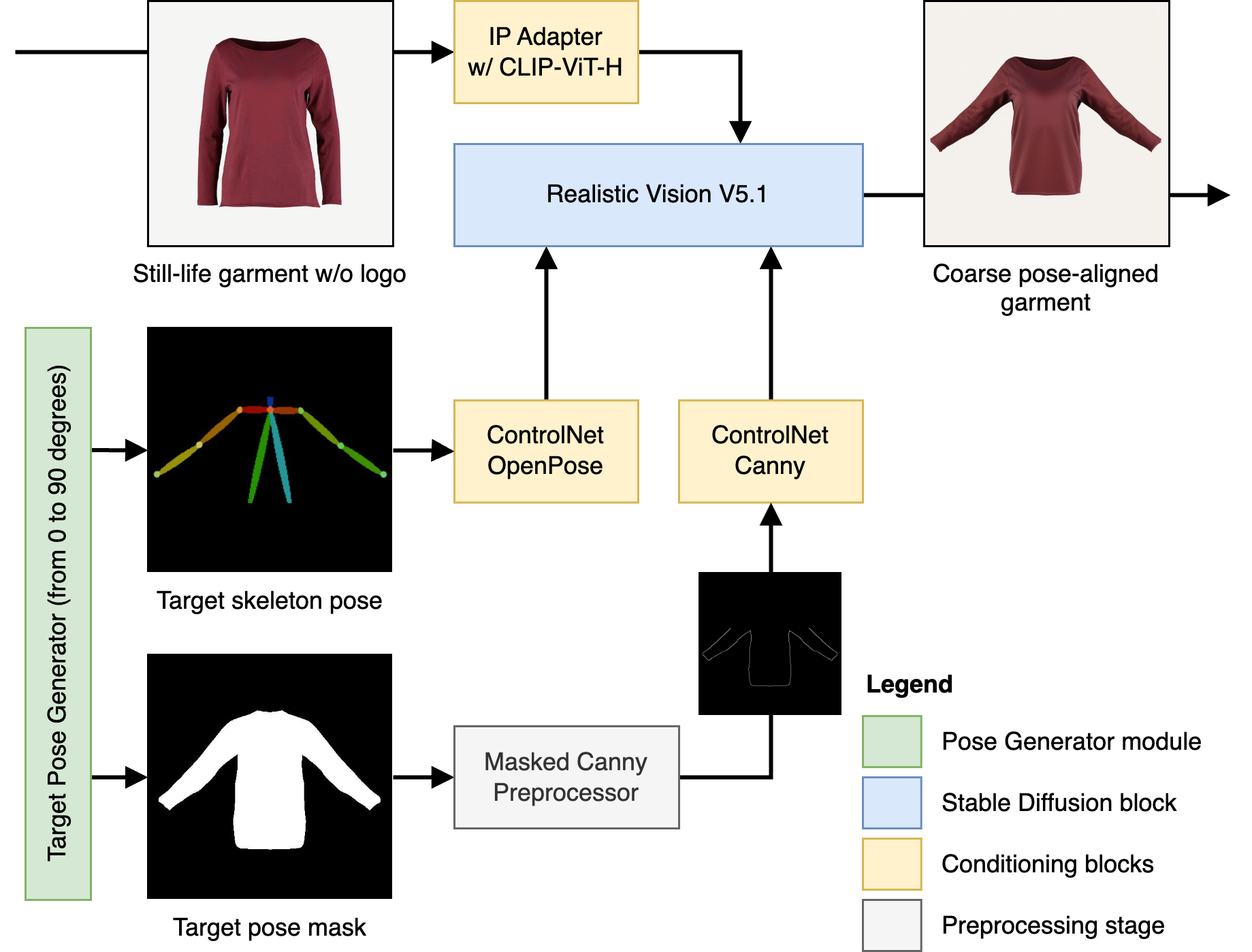}
    \caption[]{\textbf{Coarse Garment Generation.} We leverage Realistic Vision~\footnotemark\xspace and a Blender-based Pose Generator module, which generates poses with an aperture ranging from 0 to 90 degrees. This stage produces a first coarse pose-aligned clothing image starting from a still-life logo-suppressed garment. In this stage, the Stable Diffusion model is conditioned by different embeddings from both the IP-Adapter and ControlNets. The canny processed image is masked on the neck to facilitate its adaptation to diverse neckwear.}
    \label{fig:coarse_garment_generation_pipeline}
\end{figure}
\footnotetext{Realistic Vision - \url{https://civitai.com/models/4201?modelVersionId=501240}}

\subsection{Coarse Generation}
\label{sec:coarse_generation}

The Coarse Generation stage generates an initial coarse representation of a reposed garment, preserving the key details from the source image (see Fig.~\ref{fig:coarse_garment_generation_pipeline}). The process leverages state-of-the-art pretrained models, including RealisticVision~\citep{DBLP:conf/cvpr/RombachBLEO22}, ControlNets~\citep{DBLP:conf/iccv/ZhangRA23}, and IP-Adapter Plus~\citep{DBLP:journals/corr/abs-2308-06721}.

We formally define the coarse generation stage. Starting with the generation of the target pose configuration, the Target Pose Generator module, \(\mathcal{P}\), produces the target skeleton pose, \(S_{pose}\), and the corresponding pose mask, \(M_{pose}\), based on a given target aperture angle \(\theta_{target}\):

\begin{equation}
(S_{pose}, M_{pose}) = \mathcal{P}(\theta_{target}).
\end{equation}

To enhance adaptability to various neckwear designs, a binary neck mask \(M_{neck}\) is defined to filter out the neck region during edge processing. It is formulated as
\begin{equation}
(M_{neck})_{i,j} = 
\begin{cases}
1, & \text{if } i \leq 0.4 \cdot H, \\
0, & \text{otherwise},
\end{cases}
\end{equation}
where \(H\) is the image height and \((i, j)\) are pixel coordinates. The mask has the same dimensions as the input image \(I_{ls}\).

The coarse generation is obtained through a diffusion process, conditioned by various signals coming from different feature embeddings, such as \(E_{ip}\), \(E_{pose}\), and \(E_{canny}\). The image embeddings, \(E_{ip}\), are extracted from the logo-suppressed image, \(I_{ls}\), using the IP-Adapter encoder \(\mathcal{E}_{ip}\):
\begin{equation}
E_{ip} = \mathcal{E}_{ip}(I_{ls}).
\end{equation}
The pose embeddings, \(E_{pose}\), are derived from the target skeleton pose, \(S_{pose}\), using the ControlNet OpenPose encoder \(\mathcal{E}_{pose}\):
\begin{equation}
E_{pose} = \mathcal{E}_{pose}(S_{pose}).
\end{equation}
The canny embeddings, \(E_{canny}\), are computed from a masked edge map, \(\xi\), obtained by applying the \(\mathcal{X}\) edge detection operator to the pose mask \(M_{pose}\) and masking it with \((1 - M_{neck})\), representing the body area:
\begin{equation}
\xi = \mathcal{X}(M_{pose}) \odot (1 - M_{neck}), \quad E_{canny} = \mathcal{E}_{canny}(\xi).
\end{equation}
Finally, the coarse target pose image \(I_{coarse}\) is generated by a diffusion model \(\mathcal{DM}\) conditioned on all three embeddings:
\begin{equation}
I_{coarse} = \mathcal{DM}(E_{ip}, E_{pose}, E_{canny}).
\end{equation}

\begin{figure}[t]
    \centering
    \includegraphics[height=11cm]{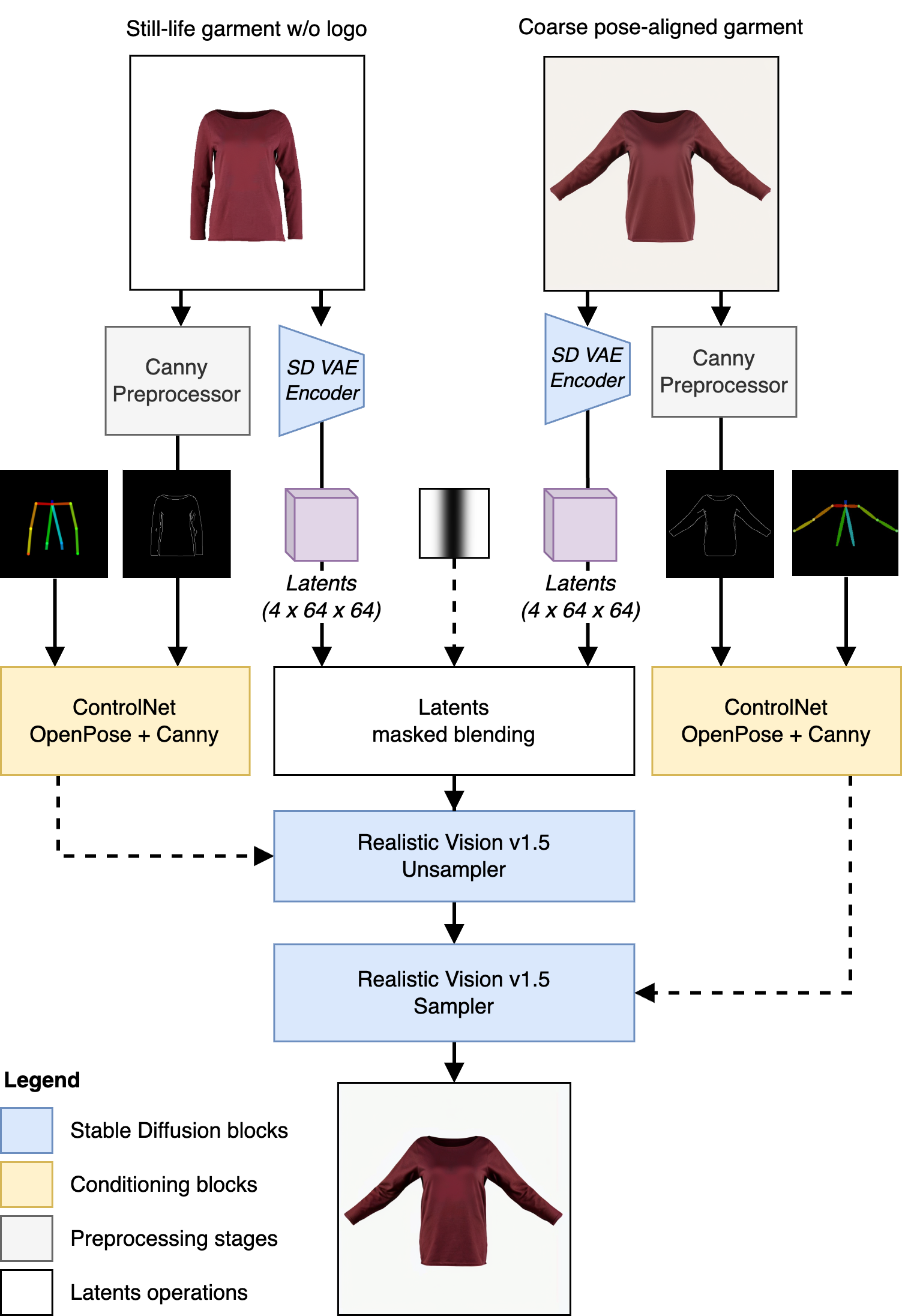}
    \caption{\textbf{Conditioned Unsampling.}  We obtain the latents and the canny-processed images from the still-life and the pose-aligned images. Then, the latents are fused through a gradient-mask blending operation, and refined through a two-pass diffusion process.}
    \label{fig:conditioned_unsampling_pipeline}
\end{figure}



%% file: 2_stage4.tex
\subsection{Conditioned Unsampling}
\label{sec:conditioned_unsampling}

The Conditioned Unsampling stage (see Fig.~\ref{fig:conditioned_unsampling_pipeline}), inspired by the recent advances in conditioned unsampling techniques~\footnote{Conditioned Unsampling - \url{https://github.com/BlenderNeko/ComfyUI_Noise}} is guided by positive and negative example embeddings. The unsampling process involves injecting conditioned noise up to a specified timestep, followed by its removal during the conditioned sampling phase. This approach facilitates navigation within the latent space, directed by embeddings, to achieve the desired sleeves transformation.

We formally define the conditioned unsampling stage, starting with latent embedding extraction. Latent embeddings are derived from both the logo-suppressed garment, \(I_{ls}\), and the coarse pose-aligned garment, \(I_{coarse}\), using the Stable Diffusion VAE encoder \(\mathcal{V}\):
\begin{equation}
Z_{ls} = \mathcal{V}(I_{ls}), \quad Z_{coarse} = \mathcal{V}(I_{coarse}).
\end{equation}
A gradient-based latent blending operation generates a fused latent representation \(Z_{blend}\):
\begin{equation}
Z_{blend} = Z_{ls} \odot Z_{gradient} + Z_{coarse} \odot (1 - Z_{gradient}),
\end{equation}
where \(Z_{gradient}\) denotes a gradient blending mask in latent space.

\noindent Skeleton poses for still-life and target configurations (given a target arm-to-torso angle \(\theta_{target}\)) are computed by the Target Pose Generator \(\mathcal{P}\):
\begin{equation}
S_{still} = \mathcal{P}(0), \quad S_{target} = \mathcal{P}(\theta_{target}).
\end{equation}

\noindent Edge maps, \(\xi_{ls}\) and \(\xi_{coarse}\), are subsequently computed by an edge detection operator \(\mathcal{X}\):
\begin{equation}
\xi_{ls} = \mathcal{X}(I_{ls}), \quad \xi_{coarse} = \mathcal{X}(I_{coarse}).
\end{equation}

\noindent The skeleton poses (\(S_{still}\) and \(S_{target}\)) and the edge maps (\(\xi_{ls}\) and \(\xi_{coarse}\)) are employed to obtain the embeddings \(\widetilde{E}_{pose}\) and \(\widetilde{E}_{canny}\) from ControlNet OpenPose, \(\mathcal{E}_{pose}\), and ControlNet Canny, \(\mathcal{E}_{canny}\), both for unsampling and the sampling process:
\begin{equation}
\widetilde{E}_{pose}^{u} = \mathcal{E}_{pose}(S_{still}), \quad \widetilde{E}_{canny}^{u} = \mathcal{E}_{canny}(\xi_{ls}),
\end{equation}
\begin{equation}
\widetilde{E}_{pose}^{s} = \mathcal{E}_{pose}(S_{target}), \quad \widetilde{E}_{canny}^{s} = \mathcal{E}_{canny}(\xi_{coarse}).
\end{equation}
These embeddings guide a two-stage diffusion process to refine the output:
\begin{equation}
Z_{unsampled} = \mathcal{DM}_{unsampler}(Z_{blend}, \widetilde{E}_{pose}^{u}, \widetilde{E}_{canny}^{u}),
\end{equation}
\begin{equation}
I_{refined} = \mathcal{DM}_{sampler}(Z_{unsampled}, \widetilde{E}_{pose}^{s}, \widetilde{E}_{canny}^{s}),
\end{equation}
where \(\mathcal{DM}_{unsampler}\) denotes the unsampling technique applied on a diffusion model and \(\mathcal{DM}_{sampler}\) represents the standard conditioned diffusion model with a given initial timestep and non-random initial latents.



%% file: 2_stage5.tex
\subsection{Source-Target Shape Matching}
\label{sec:source_target_shape_match_algorithm}

To address the misalignment between the generated reposed image and the still-life version, a shape-matching algorithm is introduced to adjust scaling and ensure precise alignment (see Fig.~\ref{fig:shape_match_comparison}). This algorithm operates on the silhouette masks of the source and target garments. It analyzes the torso section of both masks through a bounding box. Then, it extracts the heights of these bounding boxes to compute the scaling factor and the positional offset, which guides the subsequent image resizing and alignment.

Formally, this mask-guided shape alignment is defined as follows. Given the logo-suppressed garment image \(I_{ls}\), its segmentation mask \(M_{ls}\), the refined garment image \(I_{refined}\), its mask \(M_{refined}\), and the target pose mask \(M_{pose}\), the aligned images \(\overline{I}_{ls}\) and \(\overline{I}_{refined}\) are generated through:
\begin{equation}
\overline{I}_{ls} = \mathcal{M}(I_{ls}, M_{ls}, M_{pose}),
\end{equation}
\begin{equation}
\overline{I}_{refined} = \mathcal{M}(I_{refined}, M_{refined}, M_{pose}),
\end{equation}

where \(\mathcal{M}\) denotes the shape matching algorithm. This method aligns the garment images according to their silhouette masks to ensure spatial coherence, structural integrity, and precise matching of features across the transformed and original images.



%% file: 2_stage6.tex
\begin{figure}[t]
    \centering
    \begin{subfigure}{0.49\linewidth}
        \includegraphics[width=\linewidth, trim={0 2.5cm 0 2.5cm},clip]{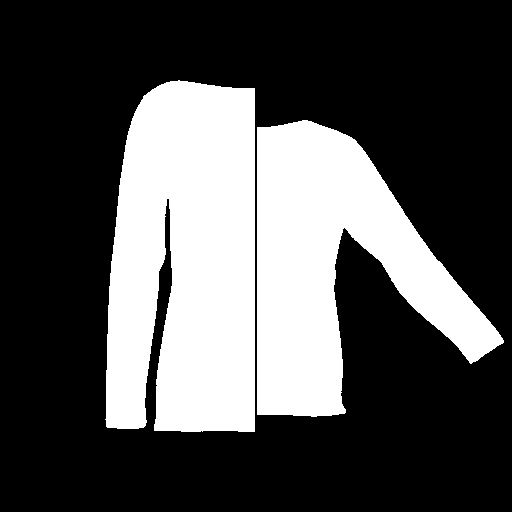}
    \end{subfigure}
    \begin{subfigure}{0.49\linewidth}
        \includegraphics[width=\linewidth, trim={0 2.5cm 0 2.5cm},clip]{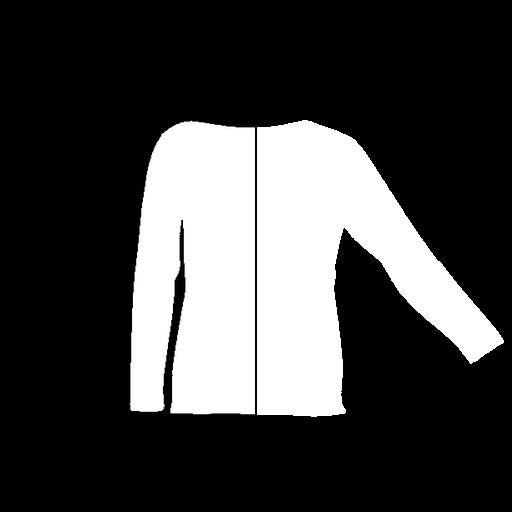}
    \end{subfigure}
    \caption{\textbf{Source-Target Shape Matching.}
    The left panel illustrates the pre-alignment comparison of the source and target garment silhouettes. Meanwhile, on the right, the silhouettes are depicted aligned after the shape matching algorithm.}
    \label{fig:shape_match_comparison}
\end{figure}

\begin{figure}[t]
    \centering
    \begin{subfigure}{0.49\linewidth}
        \includegraphics[width=\linewidth]{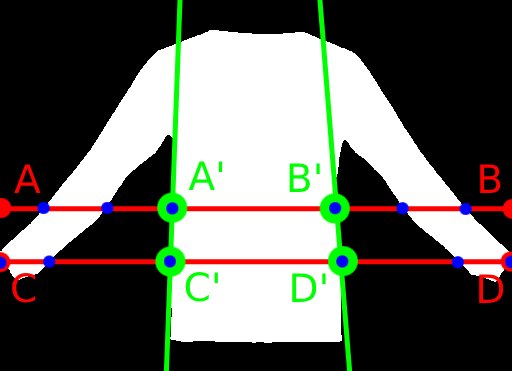}
        \label{fig:garment_parts_composition_algorithm_pipeline-a}
    \end{subfigure}
    \begin{subfigure}{0.49\linewidth}
        \includegraphics[width=\linewidth]{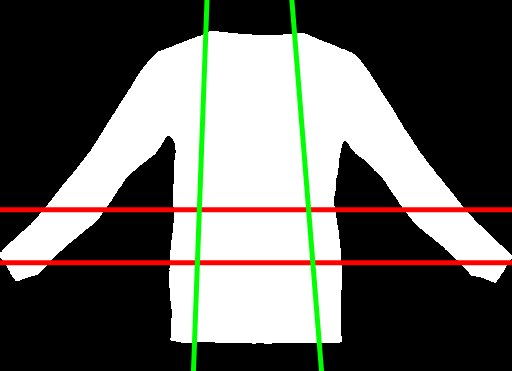}
        \label{fig:garment_parts_composition_algorithm_pipeline-a}
    \end{subfigure}
    \caption{\textbf{Garment Parts-Composition.} The figure illustrates a graphical representation of Algorithm.~\ref{alg:parts_composition_algorithm}. The left panel depicts the points, identified by the algorithm, over-imposed on the target-pose mask. Meanwhile, the right panel depicts the green lines representing the discriminating boundaries between the sleeves and the torso regions.}
    \label{fig:garment_parts_composition_algorithm_pipeline}
\end{figure}

\subsection{Garment Parts-Composition}
\label{sec:garment_parts_composition}

The garment parts-composition stage constructs the final garment by combining the still-life torso with the generated opened sleeves. This step ensures garment identity preservation and addresses potential distortions introduced in prior stages. A binary composition mask is computed to facilitate seamless merging of these elements (see Fig.~\ref{fig:garment_parts_composition_algorithm_pipeline}).

Formally, we define the garment parts-composition procedure as follows. A target pose mask at a fixed aperture of 45 degrees, \(M_{pose}^{@45}\), is initially computed:
\begin{equation}
M_{pose}^{@45} = \mathcal{P}(45),
\end{equation}
where \(\mathcal{P}\) denotes the Target Pose Generator module.

The final composed garment image, \(I_{composed}\), is then generated by the composition algorithm \(\mathcal{C}\), which merges the refined garment image \(\overline{I}_{refined}\) and the logo-suppressed still-life garment \(\overline{I}_{ls}\) guided by the computed pose mask:
\begin{equation}
I_{composed} = \mathcal{C}(\overline{I}_{refined}, \overline{I}_{ls}, M_{pose}^{@45}).
\end{equation}

To ensure robustness across varying arm positions (0 to 90 degrees), sleeve reference points are computed using the fixed 45-degree aperture mask. This ensures the overlapping region between the still-life and repositioned sleeve masks from the preceding shape-matching stage (Sec.~\ref{sec:source_target_shape_match_algorithm}) provides stable reference points for seamless merging.

The conditioned unsampling step (Sec.~\ref{sec:conditioned_unsampling}) further refines the merged boundaries, smoothing any hard transitions. Finally, a color realignment operation restores the garment's original colors, ensuring visual coherence and integrity in the final output.


\begin{algorithm}[t]
  \caption{Garment parts-composition algorithm}
  \label{alg:parts_composition_algorithm}
  \begin{algorithmic}[1]
    \State \textbf{Input:} Pose aligned mask at 45-degrees, still-life RGB clothing image, pose-aligned RGB clothing image.
    \State \textbf{Output:} Parts-composited image.
    \algrule
    \State Define points on the mask: $A(0; 0.50 \cdot h)$, $B(w; 0.50 \cdot h)$, $C(0; 0.60 \cdot h)$, and $D(w; 0.60 \cdot h)$;
    \State $r \leftarrow $ horizontal line passing through points $A$ and $B$;
    \State $t \leftarrow $ horizontal line passing through points $C$ and $D$;
    \If {$r$ and $t$ intercept 12 points on the mask}
        \State \textbf{Return} middle intersection points $A'$, $B'$, $C'$, $D'$;
    \Else
        \State \textbf{Return} a default mask;
    \EndIf
    \State $u \leftarrow $ vertical line passing through points $A'$ and $C'$;
    \State $v \leftarrow $ vertical line passing through points $B'$ and $D'$;
    \State $P \leftarrow 12$ pixels as padding factor;
    \State Shift $u$ and $v$ towards the center of the image by $P$;
    \State Consider the outer parts of the mask as sleeves and the inner part as the torso;
    \State $I' \leftarrow $ parts-composited RGB image;
    \State \textbf{Return} $I'$.
  \end{algorithmic}
\end{algorithm}

%% file: 2_stage7.tex
\subsection{Logo Injection}
\label{sec:logo_injection}

Once the garment has been reposed and all edits have been applied, the pipeline reintroduces the previously extracted logos and labels into the final image. This step ensures that brand identity is preserved, and visual coherence with the original garment design is maintained.

The logo injection process consists of two steps: upscaling and reinjection. The upscaling step adapts the composed image \(I_{composed}\) to a higher resolution using an upscaling function \(\mathcal{U}_{\uparrow}\):

\begin{equation}
I_{composed}^{\uparrow} = \mathcal{U}_{\uparrow}(I_{composed}).
\end{equation}

Subsequently, identity preservation is explicitly enforced through the reintegration of logos. The previously extracted logos \(I_{logos}\) are injected into the upscaled image \(I_{composed}^{\uparrow}\), guided by their corresponding segmentation masks \(M_{logos}\). The final output image, \(I_{out}\), is obtained as follows, 
\begin{equation}
I_{out} = \mathcal{L}^{-1}(I_{composed}^{\uparrow}, I_{logos}, M_{logos}),
\end{equation}

where \(\mathcal{L}^{-1}\) denotes the logo reinjection function, \(I_{logos}\) and \(M_{logos}\) represent the extracted logos and their associated masks, respectively. This final step guarantees that the garment's branding elements are accurately and seamlessly reincorporated into the edited garment.

%% file: 3_experiments.tex
\section{Experimental Results}
\label{sec:experiments}

This section provides an overview of the experimental setup and validates the pipeline's effectiveness through quantitative and qualitative analyses. An ablation study evaluates the impact of each pipeline stage.

\subsection{Experimental Setup}
All the experiments are conducted on a workstation equipped with an NVIDIA RTX 4090 GPU with CUDA 12.5. We are using the IP-Adapter Plus~\citep{DBLP:journals/corr/abs-2308-06721} with CLIP ViT-H/14 LAION-2B~\citep{ilharco_gabriel_2021_5143773, cherti2023reproducible, Radford2021LearningTV, schuhmann2022laionb} to enforce strong reference image conditioning. To guide the unsampler and sampler diffusion processes, we leverage a ControlNet OpenPose and a ControlNet Canny~\citep{DBLP:conf/iccv/ZhangRA23} v1.1 production models, both for positive and negative conditionings. For the cloth segmentation mask, we integrated Florence2~\citep{DBLP:conf/cvpr/0004WXDHL00Y24} and SAM2~\citep{DBLP:journals/corr/abs-2408-00714}. For the logo detection task, we utilize Florence2 and LLaMA3-8b Instruct~\citep{DBLP:journals/corr/abs-2407-21783}. We utilized RealisticVision v5.1~\citep{DBLP:conf/cvpr/RombachBLEO22} as stable diffusion model with the Karras scheduler~\citep{DBLP:conf/nips/KarrasAAL22, DBLP:conf/iclr/0011SKKEP21} and DPM++2M~\citep{DBLP:journals/corr/abs-2211-01095} as a deterministic sampler (non-SDE). Finally, for the upscale task, we leveraged the 4$\times$ Ultrasharp v1.0~\citep{DBLP:conf/eccv/WangYWGLDQL18} upscaling model. All the subsequent quantitative and qualitative evaluations are conducted with a target pose fixed at 45 degrees. In the training of the sleeve detection stage, AdamW~\citep{DBLP:conf/iclr/LoshchilovH19} and RMSprop~\footnote{Tieleman, Tijmen and Hinton, "Lecture 6.5 - rmsprop: Divide the gradient by a running average of its recent magnitude", 2012} optimizers were utilized with a batch size of 64, a learning rate of 0.0001, and a total of 70 epochs. The sleeve detection models are trained using the binary cross-entropy with logits loss, defined in Eq.~\ref{eq:binary_cross_entropy_loss_1} and Eq.~\ref{eq:binary_cross_entropy_loss_2}.
\review{The pipeline's performance relies on carefully tuned hyperparameters controlling the architecture components, such as diffusion sampling, conditioning strengths, and image processing thresholds. The hyperparameter configuration, detailed in Table~\ref{tab:hyperparameters}, was found through empirical evaluation to optimize synthesis quality and identity preservation. This calibration is essential to address the complex multi-model synchronization challenges discussed in Section~\ref{sec:integration_challenges}.}

\input{hyperparameters}

\subsection{Quantitative Evaluation}
\label{sec:quantitative_evaluation}

\input{quantitative_evaluation}

\input{qualitative_evaluation}

\input{quantitative_evaluation_ablation}

\input{qualitative_evaluation_ablation}

A detailed quantitative analysis was conducted to evaluate the efficacy of the proposed pipeline (see Table~\ref{tab:quantitative_evaluation}).

\textbf{Datasets.} The pipeline was evaluated against two different publicly available fashion datasets: DressCode~\citep{DBLP:journals/tog/HeYZYLX24} and VITON-HD~\citep{choi2021viton}, considering only images of upper body garments. We conducted an additional study on a subset containing only upper-body clothing with brand logos to better assess the effectiveness of the pipeline's logo restoration stage. To the best of our knowledge, since there are no existing datasets that specifically address garment repose, we employed datasets originally designed for Virtual Try-on tasks, \ul{even though our task is \textbf{not} a Virtual Try-on task}. In addition, we tested our pipeline on a proprietary long-sleeve garment dataset collected from the OVS catalogue, enabling evaluation with real-world products. This dataset consists of 861 long-sleeve still-life products commercialized on the OVS e-commerce platform over the past year.


\textbf{Baselines.} The pipeline was benchmarked against several baselines, including MasaCtrl~\citep{DBLP:conf/iccv/CaoWQSQZ23}, Null-text Inversion~\citep{DBLP:conf/cvpr/MokadyHAPC23}, Tuning-free Inversion-enhanced Control (TIC)~\citep{DBLP:conf/aaai/DuanCK0FFH24}, Free-Prompt-Editing (FPE)~\citep{DBLP:conf/cvpr/Liu0CJH24}, and ControlNet~\citep{DBLP:conf/iccv/ZhangRA23}. TIC and FPE architectures were re-implemented based on details provided in their respective official papers. These baselines were selected for their relevance to diffusion-based editing methods, which are prominent in the current state-of-the-art. However, each exhibits specific limitations that our method is designed to address.

\textbf{Evaluation metrics.} The baselines and our method are assessed using different quantitative metrics~\citep{5596999}, including Fréchet inception distance (FID)~\citep{DBLP:conf/nips/HeuselRUNH17}, Learned Perceptual Image Patch Similarity (LPIPS)~\citep{zhang2018perceptual}, Structural Similarity Index Measure (SSIM)~\citep{1284395}, and a domain-defined version of the Unspecified Comprehensive Affinity Measurement Index (CAMI-U)~\citep{DBLP:journals/corr/abs-2407-12705}. The metrics are defined as follows:
\begin{equation}
    \textrm{FID} = \|\mu - \mu_w\|^2 + tr(\Sigma + \Sigma_w - 2(\Sigma \Sigma_w)^{\frac{1}{2}})
\end{equation}
where \( \mu \) and \( \Sigma \) denote the mean and covariance of feature representations for the reference images, \( \mu_w \) and \( \Sigma_w \) denote the mean and covariance of the feature representations for the reposed images, and \( tr \) is the matrix trace operation.
\begin{equation}
    \textrm{LPIPS}(Y, \hat{Y}) = \sum_{l} \dfrac{1}{H_l W_l} \sum_{h, w} \| w_l \odot (x_{hw}^l - \hat{x}_{hw}^l) \|_2^2
\end{equation}
where $Y$ and $\hat{Y}$ denote the reference and the reposed images, $H_l$ and $W_l$ indicate the height and the width of the feature maps at the layer $l$, respectively. For the image pair, $x_{hw}^l$ and $\hat{x}_{hw}^l$ represent the single pixels of the source and the reposed image, while \( w_l \) is the associated weight at layer $l$. Finally, $\odot$ denotes the Hadamard product, which is used to perform element-wise multiplication between the weight vector $w_l$ and the difference of the feature maps $(x_{hw}^l - \hat{x}_{hw}^l)$. This operation allows for channel-wise scaling of the feature map differences, where each channel is weighted differently based on its perceptual relevance.
\begin{equation}
    \textrm{SSIM}(\hat{Y}, M_{pose}^{-1}) = [ l(\hat{Y}, M_{pose}^{-1})^\alpha \cdot c(\hat{Y}, M_{pose}^{-1})^\beta \cdot s(\hat{Y}, M_{pose}^{-1})^\gamma ]
\end{equation}
where $\hat{Y}$ and $M_{pose}^{-1}$ denote the reposed image and the inverted target pose mask, $l$, $c$ and $s$ are luminance, contrast, and structure factors and $\alpha$, $\beta$, $\gamma$ control the relative importance of each factor.
\begin{equation}
    S_{\textrm{CAMI-U}}(Y, \hat{Y}, M_{pose}^{-1}) = S_s(\hat{Y}, M_{pose}^{-1}) + S_t(Y, \hat{Y}) + S_k(Y, \hat{Y})
\end{equation}
where $S_{\textrm{CAMI-U}}$ denotes a modified version of the Unspecified Comprehensive Affinity Measurement Index score which accepts as input the reference image, $Y$, the reposed image, $\hat{Y}$, and the inverted target pose mask, $M_{pose}^{-1}$. Finally, $S_s$, $S_t$, and $S_k$ represent the scores for the image structure, texture and keypoints, respectively.

In the absence of a suitable ground truth for directly assessing the reposed images, the evaluation is conducted by calculating the FID and LPIPS metrics between the reference and generated images, specifically constraining the analysis to the torso region defined by the composition mask derived from Algorithm~\ref{sec:source_target_shape_match_algorithm}. FID and LPIPS measures are employed to quantify perceptual similarity within this masked region. To assess the structural integrity of the reposed image, the SSIM is computed by comparing the full output image with the inverted alpha mask generated by the Pose Generator module. Following the same setup, we evaluated the pipeline with CAMI-U score by computing the image structure score between the reposed image and the inverted alpha mask of the target pose. In contrast, the texture similarity score and keypoints matching score are computed considering the full reference and reposed images.

Given that our method uniquely addresses the garment repose problem, it demonstrates robust performance in terms of SSIM, surpassing existing methodologies.
Furthermore, our approach achieves superior results in CAMI-U across nearly all evaluated scenarios, and outperforms state-of-the-art methods in FID when evaluated on DressCode and VITON-HD subsets consisting exclusively of garments featuring brand logos. Across the remaining benchmarks, the performances remain competitive. To facilitate the interpretation of these results, a visualization through a Pareto front is provided in Figure~\ref{fig:pareto_quantitative_evaluation}.

\input{pareto_sota}

To select the best model for the Long Sleeve Detection stage, eight different experiments were conducted. The quantitative results of these experiments are summarized in Table~\ref{tab:sleeve_detection_results}, which provides an overview of the performance of the models in terms of accuracy, precision, recall, and F1 score. Considering the results obtained, we selected the VGG16 model trained with RMSprop optimizer as the best performing model. It presents the highest accuracy, precision, recall, and F1 score among all the evaluated models.

To demonstrate the computational feasibility of our method in a production scenario, we present a runtime analysis in Table~\ref{tab:runtime_analysis} for the OVS dataset, considering execution times and GPU video memory (VRAM) consumption across all baselines. Finally, we showcase the VRAM utilization over time for our method in Figure~\ref{fig:runtime_analysis_vram_usage}.

\input{runtime_analysis}

\input{vram_usage_ours}


\subsection{Qualitative Evaluation}
\label{sec:qualitative_evaluation}

A qualitative comparison is presented in Figure~\ref{fig:qualitative_evaluation}. Our method is the only one capable of solving the pose alignment task, giving a still-life image of the garment. The presented examples demonstrate the effectiveness of our approach in preserving garment identity, texture fidelity, and branding attributes while transitioning garments from their initial poses to novel pose configurations. By contrast, the baseline methods exhibit significant shortcomings, such as distortions in sleeve shape, texture inconsistencies, and color shifts, with these issues being particularly pronounced in the outputs of MasaCtrl and Null-Text Inversion. While TIC and FPE achieves partial success by reconstructing the original texture, they fail to reposition the sleeves effectively, underscoring its limitations compared to our approach. Furthermore, our pipeline demonstrates superior performance when applied to garments featuring logos.
\review{To further strengthen our findings and broaden the applicability of FashionRepose, we conduct an additional exploratory qualitative assessment using cherry-picked samples from FashionTryOn~\citep{DBLP:conf/mm/ZhengSCHCN19} dataset. This dataset, developed explicitly for virtual try-on scenarios, provides a valuable proxy for evaluating model generalization. As shown in Figure.~\ref{fig:qualitative_evaluation_fashiontryon}, our method maintains stable identity preservation and pose alignment even if garments exhibit different style characteristics. 
However, we observed a higher occurrence of artifacts--such as blurred textures and lack of fine details--appearing more frequently in this dataset.
Upon inspection, these issues are primarily attributable to the inherently low resolution of the original images, limiting the available structural and textural information for any downstream task. Despite these constraints, FashionRepose still produces outputs preserving garment semantics and visual coherence, highlighting the pipeline's robustness even in lower-resolution visual domains.}
The results highlight that our method not only performs precise pose alignment but also handles logo suppression and reintegration with considerable fidelity, ensuring that brand identity remains intact throughout the transformation process.

\input{qualitative_evaluation_fashiontryon}

\subsection{Ablation Study}
\label{sec:ablation_study}

An ablation study was performed to analyze the contributions of individual stages within the pipeline. Table~\ref{tab:quantitative_evaluation_ablation} presents the results for each evaluation metric. The incremental incorporation of stages into the pipeline demonstrates a consistent improvement in performance metrics up to the part composition stage. However, a slight decline in performance is observed between the part composition and logo restoration stages, attributed to additional noise introduced during the final operation. The datasets, which lack a significant number of garments with brand logos, negatively influence the performance metrics of the logo restoration stage. To address this limitation, the analysis was extended to a subset containing only images of garments with brand logos. This focused subset yielded improved quantitative and qualitative results, validating the pipeline’s effectiveness in handling logo restoration tasks (see Fig.~\ref{fig:qualitative_evaluation_ablation}). To facilitate the interpretation of these results, a visualization through a Pareto front is provided in Figure~\ref{fig:pareto_ablation_evaluation}.

\input{pareto_ablation}

\subsection{Discussion}

The results of our experiments provide compelling evidence of the efficacy and robustness of the \textsc{FashionRepose} pipeline across a variety of use cases, both in controlled benchmarks and real-world industrial applications. From the outset, \textsc{FashionRepose} has demonstrated its strength in combining precision, speed, and brand fidelity (core demands of the fashion industry) without relying on fine-tuning or additional training data.

Quantitatively, our pipeline outperformed or remained competitive with state-of-the-art diffusion-based editing techniques across all major metrics, including FID, LPIPS, SSIM, and CAMI-U. Particularly notable is its consistent superiority in the sleeve repose task, evidenced by the high SSIM scores across datasets and its leading performance on the CAMI-U index, which evaluates structure, texture, and keypoints integrity. On subsets featuring branded garments, \textsc{FashionRepose} clearly excelled, achieving the lowest FID scores and strongest identity retention, validating the effectiveness of our logo restoration module. These results are especially critical given the importance of brand consistency in fashion imagery.

Qualitative results further reinforce these findings. As illustrated in our visual comparisons, \textsc{FashionRepose} is the only approach among those tested that can perform realistic pose alignment on still-life fashion images while preserving garment structure, texture, and logos. Competing methods struggled with one or more of these dimensions, often producing distorted sleeves, misaligned features, or color shifts. Our method maintained clean transitions, accurate repositioning, and visually coherent outputs, particularly in the presence of logos.

Our ablation study sheds light on the contributions of each pipeline component. The coarse generation stage laid the foundation for pose alignment, but it was the successive stages -- particularly conditioned unsampling and parts composition -- that added substantial fidelity and realism. The final logo restoration step introduced slight metric fluctuations due to minor reinjection noise, but a focused evaluation on branded garments demonstrated its tangible value in preserving commercial identity. This modular structure also highlights the pipeline’s interpretability and debuggability, an advantage in industrial contexts.

The pipeline proved its computational feasibility for production environments. When tested on OVS's proprietary dataset of 861 long-sleeve garments, \textsc{FashionRepose} achieved real-time performance with reasonable GPU memory usage and execution times, confirming its practical deployability at scale. The runtime analysis showcased how our pipeline strikes a strong balance between quality and efficiency -- outperforming several baselines in resource optimization without compromising on output integrity.

Overall, these outcomes illustrate how \textsc{FashionRepose} meets the unique demands of fashion image editing, ensuring identity-preserving, high-fidelity pose transformations in a zero-shot and training-free manner. This positions it not only as a research contribution but as a real-world solution, already validated in commercial workflows and adaptable to the fast-evolving needs of digital fashion.

\subsection{Architectural Design Challenges}
\label{sec:integration_challenges}


\review{The efficacy of \textsc{FashionRepose} depends on the individual capabilities of its constituent pretrained models and, more critically, on the sophisticated architecture devised to combine their operations. Integrating diverse modules with different input resolutions, conditioning mechanisms, and output characteristics into a cohesive, training-free architecture presents substantial challenges. Our approach carefully addresses these, particularly concerning resolution mismatches, the delicate balance of conditioning strengths, the precision of pose guidance, and the seamless management of intermediate data flows (for more information see Algorithm~\ref{alg:pose_norm_pipeline}).
}

\review{A primary limitation in such a composite system is the management of image resolutions. Pretrained diffusion models exhibit optimal performance and computational efficiency at specific resolutions, often $512 \times 512$ pixels. However, the demands of fashion image editing, especially the preservation of fine textural details and brand-specific logos, necessitate higher fidelity. To reconcile these conflicting requirements, \textsc{FashionRepose} implements a strategic multi-resolution workflow. The core generative and editing stages, including Coarse Generation (Sec.~\ref{sec:coarse_generation}) and Conditioned Unsampling (Sec.~\ref{sec:conditioned_unsampling}), are executed at a $512 \times 512$ resolution. This choice was an empirically validated trade-off, providing sufficient detail for robust garment shape and pose manipulation while maintaining manageable processing times suitable for near real-time industrial applications. For the critical logo restoration phase, however, the pipeline transitions to a higher resolution of $1024 \times 1024$ pixels, also serving as the final target output resolution. We extract the original logo and its corresponding mask from the input image, rescaled to $1024 \times 1024$ pixels. After the garment pose is altered and upscaled to $1024 \times 1024$, the previously extracted logo is reinjected. This solution circumvents the quality degradation that would inevitably occur if the logo were downscaled and subsequently upscaled, ensuring that brand elements retain their sharpness and integrity, a non-negotiable aspect for high-quality industry standards. This dual-resolution strategy effectively isolates the high-fidelity requirements of branding from the primary generative tasks, optimizing quality and efficiency.}


\review{Balancing the strength of multiple conditioning signals is another challenging aspect. The proposed pipeline leverages IP-Adapter for image-based conditioning and ControlNets (OpenPose and Canny) for structural guidance. The relative influence of these conditions must be carefully calibrated: an over-reliance on one of them may suppress the desired effects of others or introduce artifacts. This aspect is particularly pertinent in the Conditioned Unsampling stage, where the interplay between the still-life and target pose latents, guided by their respective Canny and OpenPose embeddings, defines the transformation's accuracy. Through extensive empirical tuning of conditioning scale factors and guidance strengths for each ControlNet and IP-Adapter, we found sets of hyperparameters -- showcased in Table~\ref{tab:hyperparameters} -- that harmoniously blend these influences, ensuring the garment identity is preserved while accurately conforming to the target pose without undesirable distortions. This iterative hyperparameter optimization process represents a key engineering trade-off, balancing the need for perfect control against the practicalities of finding a generalizable set of parameters.}

\review{Finally, the novel Blender-based Pose Generator module plays an essential role in our architecture by providing precise and consistent target skeleton poses and alpha masks. These high-quality, generated conditioning images for the ControlNet modules ensure that the generated pose aligns accurately with the desired output and that the segmentation masks for subsequent stages are reliable. These engineering decisions, often involving iterative refinement and adjustments based on domain experts' observations, are essential for the pipeline's ability to deliver coherent, high-quality results by effectively orchestrating the contributions of diverse, specialized modules.}

%% file: hyperparameters.tex
\begin{table}[t]
    \review{
    \centering
    \scriptsize
    \rowcolors{2}{}{gray!20}
    \begin{tabular}{@{}p{.2\linewidth}p{.35\linewidth}p{.35\linewidth}@{}}
      \toprule
      \textbf{Submodule} & \textbf{Hyperparameter} & \textbf{Optimal value} \\
      \midrule
      \rowcolor{white} \multicolumn{3}{c}{\textit{Logo Detection and Suppression}} \\
      \midrule
      Logo Detection & Grow Mask & 10 pixels \\
      Suppression & Inference Steps & 10 \\
      Suppression & Classifier-Free Guidance & 1.5 \\
      \midrule
      \rowcolor{white} \multicolumn{3}{c}{\textit{Coarse Generation}} \\
      \midrule
      Sampling & Inference Steps & 30 \\
      Sampling & Classifier-Free Guidance & 3 \\
      Conditioning & IP-Adapter Weight & 1 \\
      \midrule
      \rowcolor{white} \multicolumn{3}{c}{\textit{Conditioned Unsampling}} \\
      \midrule
      Edge Detection & Canny Edge Thresholds & Low: 100, High: 200 \\
      Conditioning & IP-Adapter Weight & 1 \\
      Conditioning & ControlNet Strengths & OpenPose: 0.60, Canny: 0.60 \\
      Mask Blending & Thresholds & Left stop: 35\%, Right stop: 65\% \\
      Mask Blending & Blur Radius & 200 pixels \\
      Unsampler & Steps & 30 \\
      Unsampler & Classifier-Free Guidance & 1 \\
      Unsampler & Timestep Stop & 0 \\
      Sampler & Timestep Start & 0 \\
      Sampler & Steps & 30 \\
      Sampler & Classifier-Free Guidance & 3 \\
      \bottomrule
    \end{tabular}
    \caption{\review{\textbf{Hyperparameters.} Structured overview of the adopted hyperparameters across different stages of the pipeline.}}
    \label{tab:hyperparameters}
    }
\end{table}


%% file: quantitative_evaluation.tex
\begin{table}[!h]
    \centering
    \scriptsize
    \setlength{\tabcolsep}{4pt}
    \rowcolors{2}{}{gray!20}
    \begin{tabular}{@{}lccccccc@{}}
      \toprule
      \textbf{Method} & \textbf{FID} ($\downarrow$) & \multicolumn{2}{c}{\textbf{LPIPS} ($\downarrow$)} & \multicolumn{2}{c}{\textbf{SSIM} ($\uparrow$)} & \multicolumn{2}{c}{\textbf{CAMI-U} ($\uparrow$)} \\
      \cmidrule(r){3-4} \cmidrule(r){5-6} \cmidrule(r){7-8}
      \rowcolor{white} & & \textbf{Mean} & \textbf{Std} & \textbf{Mean} & \textbf{Std} & \textbf{Mean} & \textbf{Std} \\
      \midrule
      \rowcolor{white} \multicolumn{8}{c}{\textit{DressCode (upper-body)}} \\
      \midrule
      MasaCtrl \citep{DBLP:conf/iccv/CaoWQSQZ23} & \textbf{9.2622} & \underline{0.0479} & \underline{0.0241} & 0.5424 & 0.0615 & 2.4682 & 0.3914 \\
      Null-text Inv. \citep{DBLP:conf/cvpr/MokadyHAPC23} & 23.0075 & 0.084 & 0.0288 & 0.5266 & 0.0741 & 2.3361 & 0.4310 \\
      TIC \citep{DBLP:conf/aaai/DuanCK0FFH24} & \underline{10.7059} & 0.0508 & 0.0269 & \underline{0.5463} & \underline{0.0658} & \underline{2.4895} & \underline{0.3509} \\
      FPE \citep{DBLP:conf/cvpr/Liu0CJH24} & 12.2550 & \textbf{0.0451} & \textbf{0.0187} & 0.5458 & 0.0571 & 2.4829 & 0.3433 \\
      ControlNet \citep{DBLP:conf/iccv/ZhangRA23} & 63.8593 & 0.0935 & 0.0225 & 0.2534 & 0.2113 & 2.3296 & 0.1760 \\
      Ours & 13.4888 & 0.0624 & 0.0186 & \textbf{0.7708} & \textbf{0.1177} & \textbf{2.5215} & \textbf{0.3269} \\
      \midrule
      \rowcolor{white} \multicolumn{8}{c}{\textit{DressCode (upper-body with brand logos)}} \\
      \midrule
      MasaCtrl \citep{DBLP:conf/iccv/CaoWQSQZ23} & 54.0739 & 0.0473 & 0.0203 & 0.5599 & 0.0520 & 2.5913 & 0.2277 \\
      Null-text Inv. \citep{DBLP:conf/cvpr/MokadyHAPC23} & 84.0337 & 0.0809 & 0.0303 & 0.5387 & 0.0712 & 2.4023 & 0.3789 \\
      TIC \citep{DBLP:conf/aaai/DuanCK0FFH24} & 46.5414 & \underline{0.0438} & \underline{0.0221} & \underline{0.5636} & \underline{0.0526} & \underline{2.6084} & \underline{0.1722} \\
      FPE \citep{DBLP:conf/cvpr/Liu0CJH24} & \underline{43.0954} & \textbf{0.0383} & \textbf{0.0149} & 0.5625 & 0.0503 & 2.6054 & 0.1902 \\
      ControlNet \citep{DBLP:conf/iccv/ZhangRA23} & 105.4250 & 0.0896 & 0.0185 & 0.3049 & 0.1951 & 2.3902 & 0.1434 \\
      Ours & \textbf{42.3898} & 0.0573 & 0.0163 & \textbf{0.8037} & \textbf{0.1242} & \textbf{2.6269} & \textbf{0.2011} \\
      \midrule
      \rowcolor{white} \multicolumn{8}{c}{\textit{VITON-HD}} \\
      \midrule
      MasaCtrl \citep{DBLP:conf/iccv/CaoWQSQZ23} & 10.7364 & \underline{0.0385} & \underline{0.0187} & 0.5283 & 0.0742 & 2.4408 & 0.4048 \\
      Null-text Inv. \citep{DBLP:conf/cvpr/MokadyHAPC23} & 21.2574 & 0.0679 & 0.0278 & 0.5155 & 0.0988 & 2.3213 & 0.4365 \\
      TIC \citep{DBLP:conf/aaai/DuanCK0FFH24} & \textbf{9.3597} & \underline{0.0385} & \underline{0.0186} & 0.5294 & 0.0837 & \underline{2.4594} & \underline{0.3798} \\
      FPE \citep{DBLP:conf/cvpr/Liu0CJH24} & \underline{10.7925} & \textbf{0.0384} & \textbf{0.0181} & \underline{0.5341} & \underline{0.0719} & 2.4506 & 0.3821 \\
      ControlNet \citep{DBLP:conf/iccv/ZhangRA23} & 80.1457 & 0.0950 & 0.0241 & 0.2374 & 0.1892 & 2.3301 & 0.1827 \\
      Ours & 20.2046 & 0.0627 & 0.0175 & \textbf{0.8113} & \textbf{0.1135} & \textbf{2.4974} & \textbf{0.3505} \\
      \midrule
      \rowcolor{white} \multicolumn{8}{c}{\textit{VITON-HD (garments with brand logos)}} \\
      \midrule
      MasaCtrl \citep{DBLP:conf/iccv/CaoWQSQZ23} & 132.5054 & 0.0512 & 0.0192 & 0.5057 & 0.1430 & 2.6097 & 0.1485 \\
      Null-text Inv. \citep{DBLP:conf/cvpr/MokadyHAPC23} & 176.8526 & 0.0740 & 0.0263 & 0.5118 & 0.0793 & 2.4961 & 0.2719 \\
      TIC \citep{DBLP:conf/aaai/DuanCK0FFH24} & 118.9420 & \underline{0.0474} & \underline{0.0192} & \underline{0.5252} & \underline{0.1008} & 2.6210 & 0.0510 \\
      FPE \citep{DBLP:conf/cvpr/Liu0CJH24} & \underline{81.0869} & \textbf{0.0406} & \textbf{0.0220} & 0.5146 & 0.1187 & \underline{2.6400} & \underline{0.0546} \\
      ControlNet \citep{DBLP:conf/iccv/ZhangRA23} & 190.2084 & 0.0866 & 0.0125 & 0.1955 & 0.1947 & 2.4177 & 0.0742 \\
      Ours & \textbf{68.8975} & 0.0510 & 0.0150 & \textbf{0.7760} & \textbf{0.1352} & \textbf{2.6949} & \textbf{0.0472} \\
      \midrule
      \rowcolor{white} \multicolumn{8}{c}{\textit{OVS (long sleeve garments)}} \\
      \midrule
      MasaCtrl \citep{DBLP:conf/iccv/CaoWQSQZ23} & 45.2648 & 0.0580 & 0.0180 & 0.5332 & 0.0402 & 2.4938 & 0.3712 \\
      Null-text Inv. \citep{DBLP:conf/cvpr/MokadyHAPC23} & 70.0711 & 0.0845 & 0.0307 & 0.4994 & 0.0611 & 2.2995 & 0.4623 \\
      TIC \citep{DBLP:conf/aaai/DuanCK0FFH24} & \textbf{23.0295} & \textbf{0.0381} & \textbf{0.0180} & 0.5342 & 0.0516 & \textbf{2.5589} & \textbf{0.2960} \\
      FPE \citep{DBLP:conf/cvpr/Liu0CJH24} & 31.9903 & \underline{0.0402} & \underline{0.0172} & \underline{0.5356} & \underline{0.0477} & 2.5170 & 0.3348 \\
      ControlNet \citep{DBLP:conf/iccv/ZhangRA23} & 108.4469 & 0.0978 & 0.0195 & 0.4515 & 0.0936 & 2.2999 & 0.2188 \\
      Ours & \underline{30.3685} & 0.0563 & 0.0139 & \textbf{0.7390} & \textbf{0.1094} & \underline{2.5588} & \underline{0.2940} \\
      \bottomrule
    \end{tabular}
    \caption{\textbf{Quantitative Evaluation.} Comparison of the baselines on DressCode \citep{DBLP:journals/tog/HeYZYLX24} and VITON-HD \citep{choi2021viton} with the corresponding subsets containing only garments with brand logos. An additional analysis was conducted on the production dataset of OVS. Best results are reported in \textbf{bold}, while second-best are \underline{underlined}.}
    \label{tab:quantitative_evaluation}
\end{table}

%% file: qualitative_evaluation.tex
\begin{figure*}[t]
  \centering
  \resizebox{\textwidth}{!}{
      \includegraphics[height=2cm]{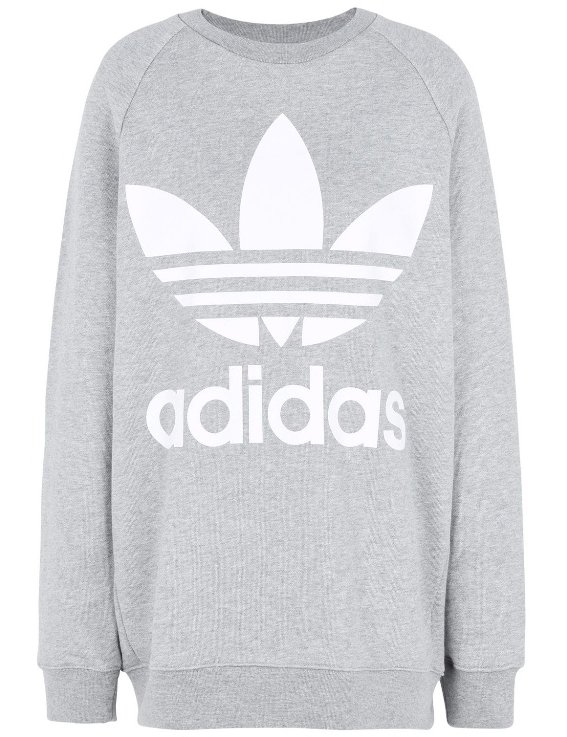}
      \includegraphics[height=2cm]{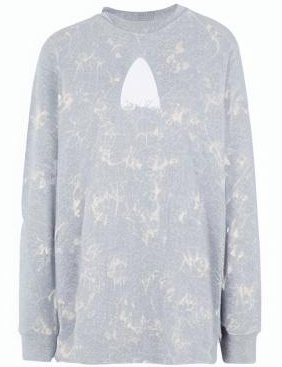}
      \includegraphics[height=2cm]{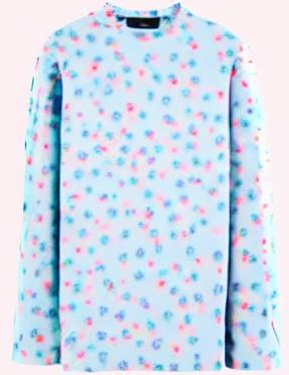}
      \includegraphics[height=2cm]{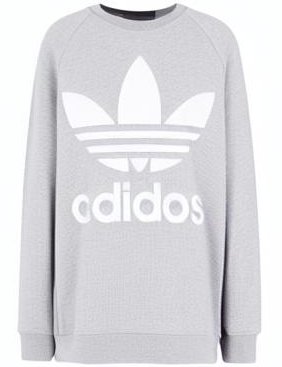}
      \includegraphics[height=2cm]{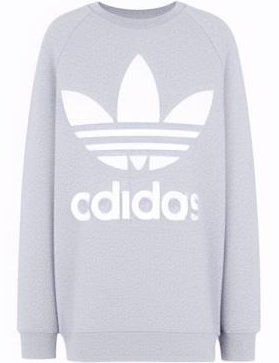}
      \includegraphics[height=2cm]{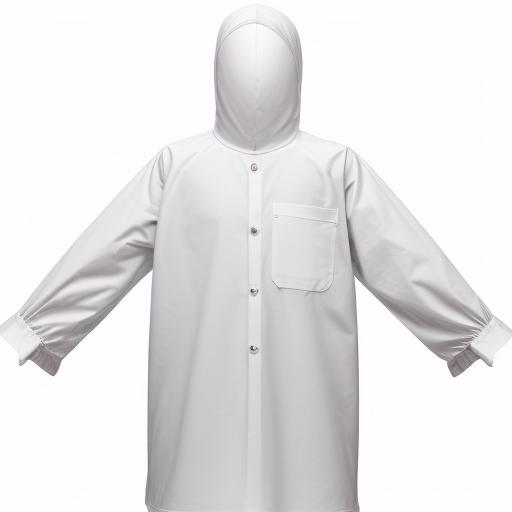}
      \includegraphics[height=2cm]{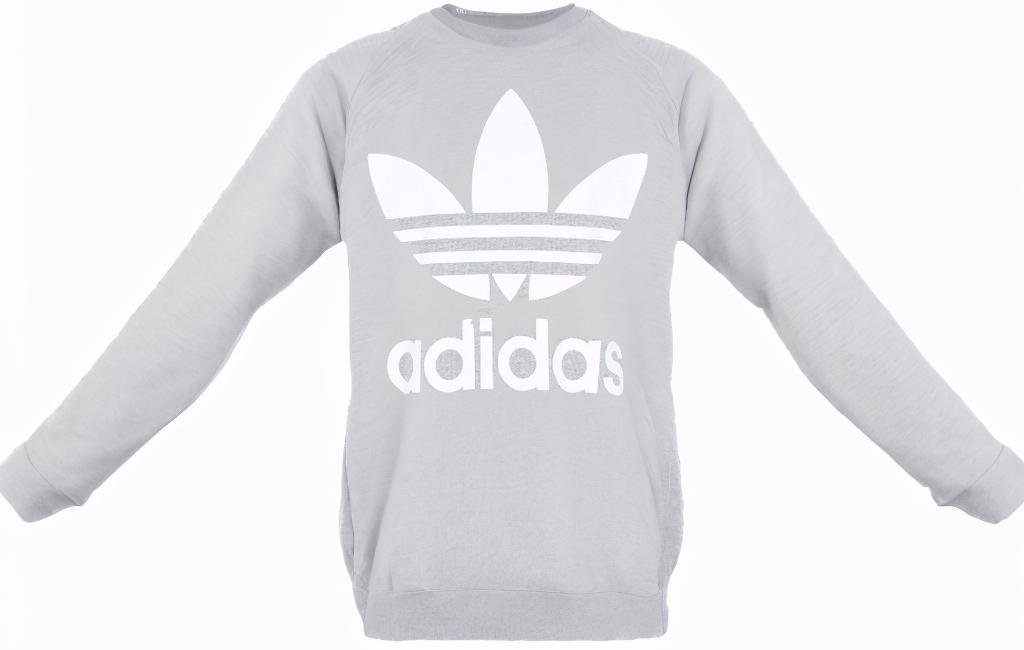}
  } \\
  \resizebox{\textwidth}{!}{
      \includegraphics[height=2cm]{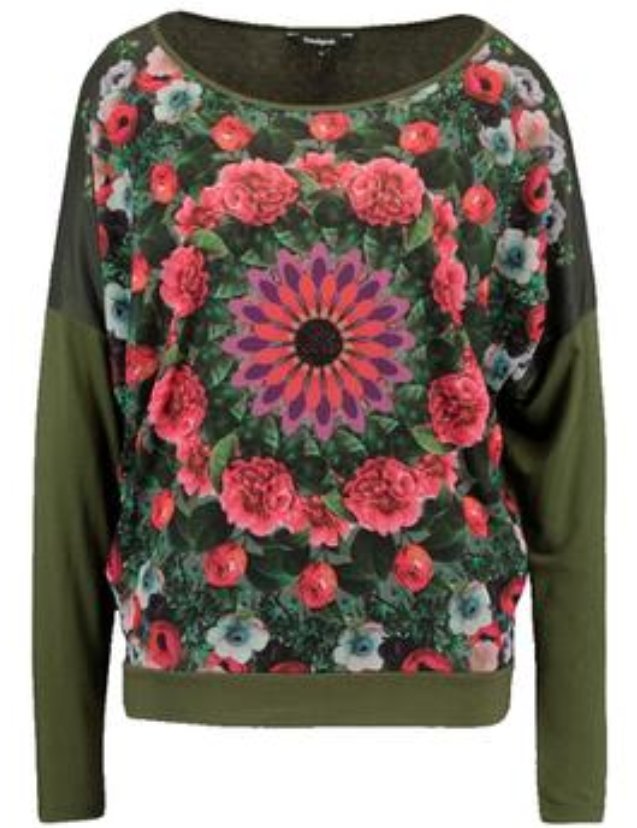}
      \includegraphics[height=2cm]{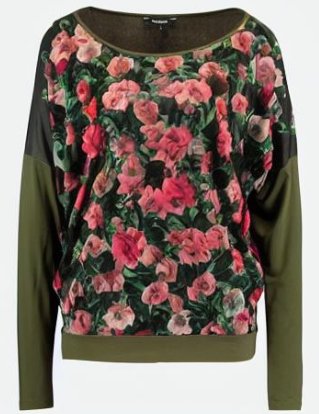}
      \includegraphics[height=2cm]{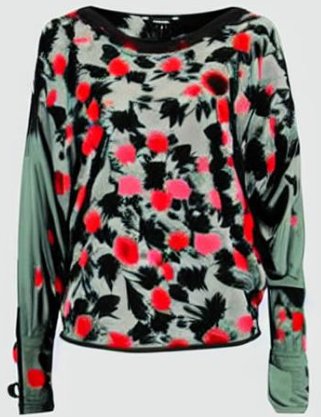}
      \includegraphics[height=2cm]{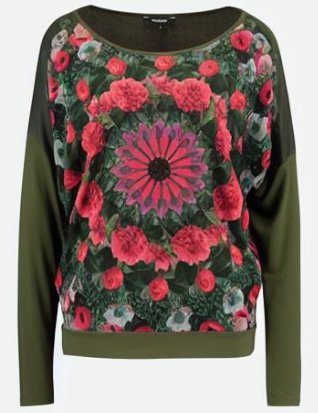}
      \includegraphics[height=2cm]{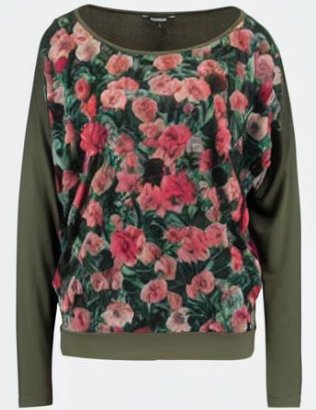}
      \includegraphics[height=2cm]{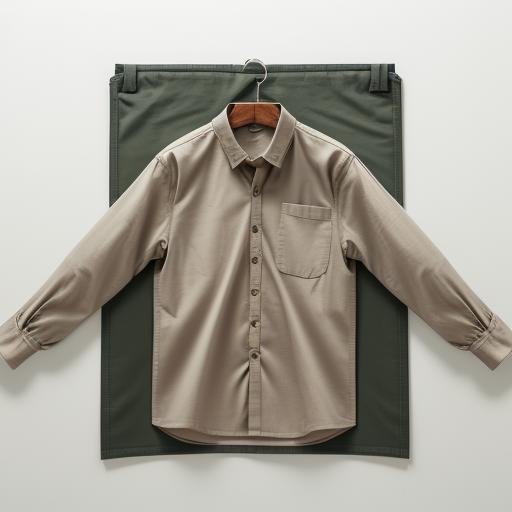}
      \includegraphics[height=2cm]{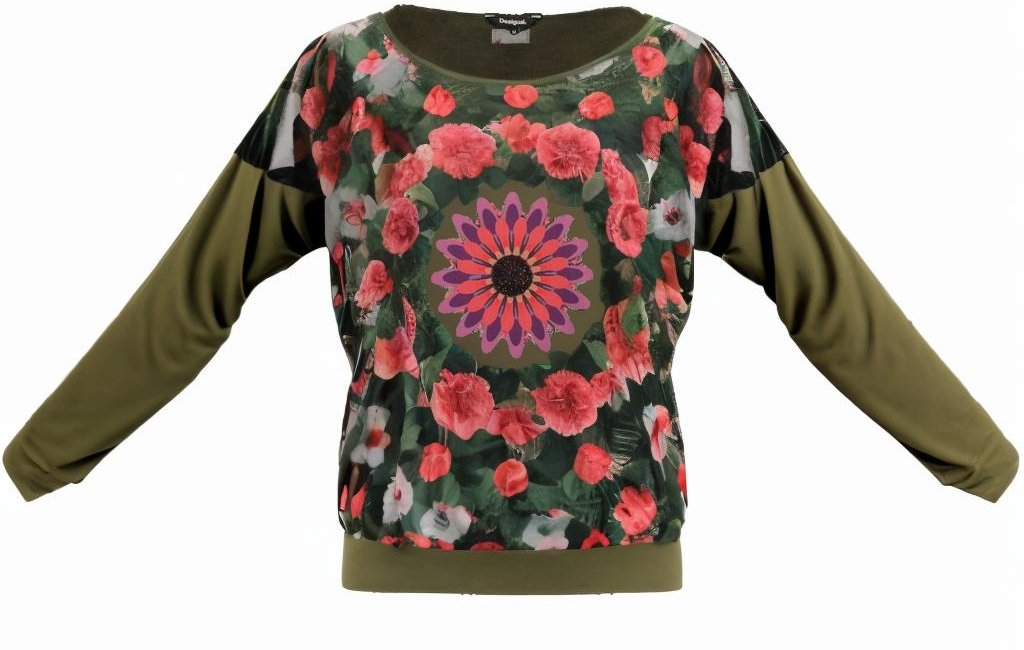}
  } \\
  \resizebox{\textwidth}{!}{
      \includegraphics[height=2cm]{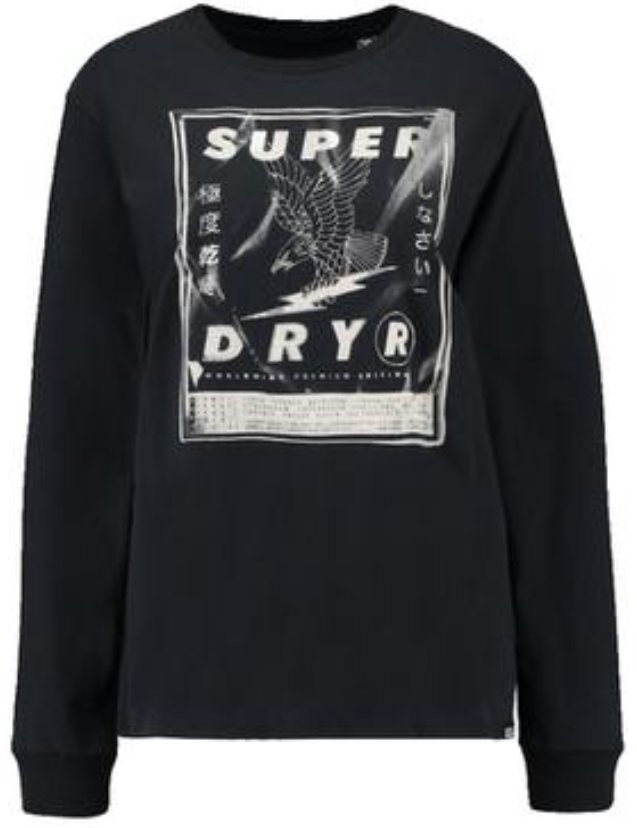}
      \includegraphics[height=2cm]{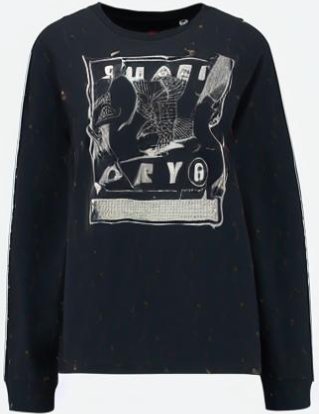}
      \includegraphics[height=2cm]{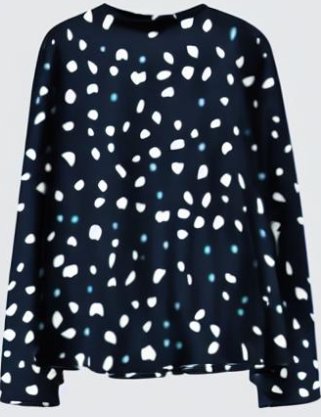}
      \includegraphics[height=2cm]{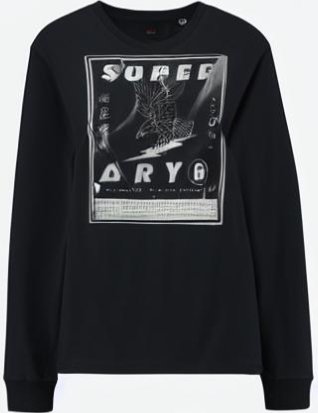}
      \includegraphics[height=2cm]{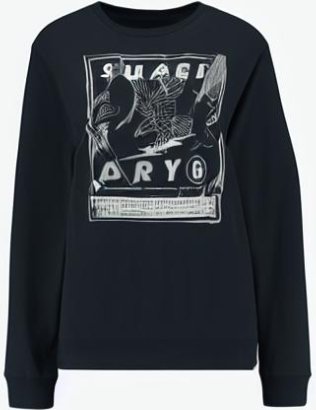}
      \includegraphics[height=2cm]{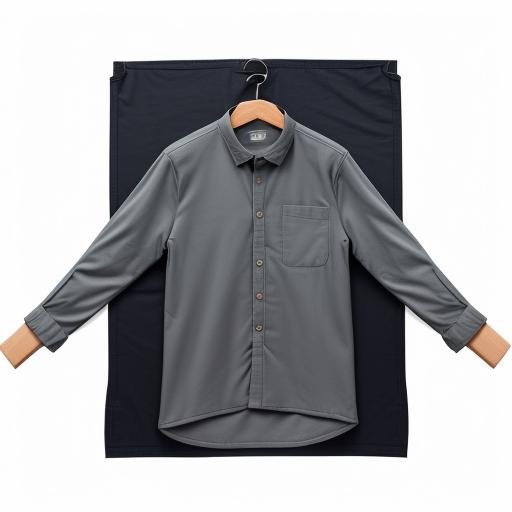}
      \includegraphics[height=2cm]{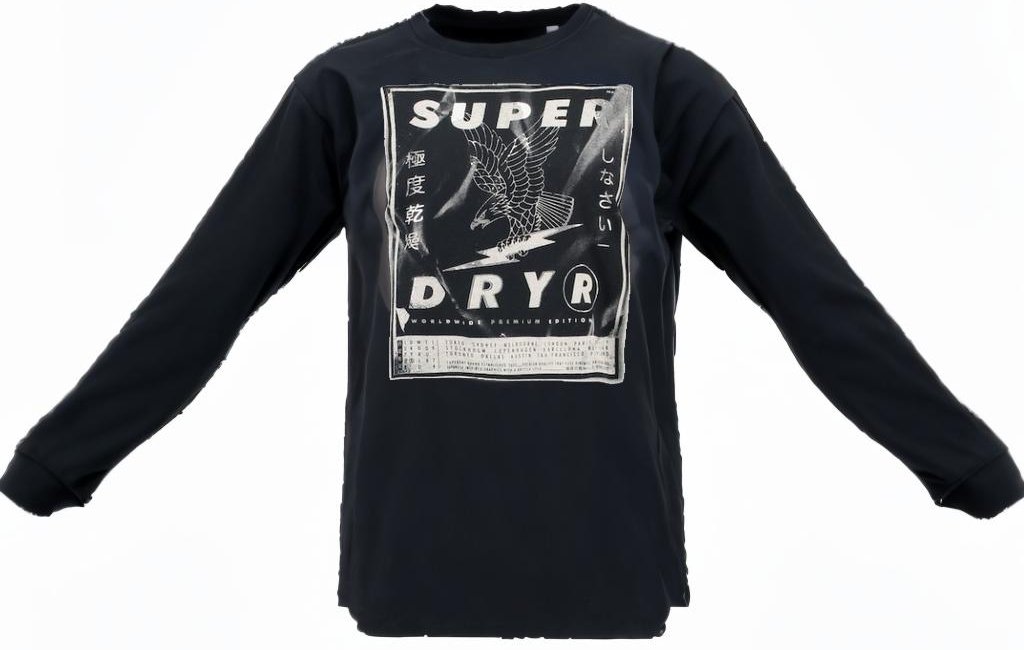}
  } \\
  \begin{tabular}{@{}c@{}c@{}c@{}c@{}c@{}c@{}c@{}}
      \subcaptionbox{Ref.}[0.12\textwidth]{}
      & \subcaptionbox{Masactrl}[0.12\textwidth]{}
      & \subcaptionbox{NullText}[0.12\textwidth]{}
      & \subcaptionbox{TIC}[0.12\textwidth]{}
      & \subcaptionbox{FPE}[0.12\textwidth]{}
      & \subcaptionbox{ControlNet}[0.15\textwidth]{}
      & \subcaptionbox{Ours}[0.25\textwidth]{}
  \end{tabular}
  \caption{\textbf{Qualitative Evaluation.} The figure illustrates the qualitative evaluation of different baselines against different clothing images sampled from DressCode~\citep{DBLP:journals/tog/HeYZYLX24} (first row) and VITON-HD~\citep{choi2021viton} (second and third row).}
  \label{fig:qualitative_evaluation}
\end{figure*}

%% file: quantitative_evaluation_ablation.tex
\begin{table}[!h]
  \centering
  \footnotesize
  \setlength{\tabcolsep}{4pt}
  \rowcolors{2}{}{gray!20}
  \begin{tabular}{@{}lcccccccc@{}}
    \toprule
    \textbf{Pipeline stage} & \textbf{FID} ($\downarrow$) & \multicolumn{2}{c}{\textbf{LPIPS} ($\downarrow$)} & \multicolumn{2}{c}{\textbf{SSIM} ($\uparrow$)} & \multicolumn{2}{c}{\textbf{CAMI-U} ($\uparrow$)} \\
    \cmidrule(r){3-4} \cmidrule(r){5-6} \cmidrule(r){7-8}
    \rowcolor{white} & & \textbf{Mean} & \textbf{Std} & \textbf{Mean} & \textbf{Std} & \textbf{Mean} & \textbf{Std} \\
    \midrule
    \rowcolor{white} \multicolumn{8}{c}{\textit{DressCode (upper-body)}} \\
    \midrule
    Coarse generation & 24.0558 & 0.0749 & 0.0211 & \textbf{0.7956} & \textbf{0.1029} & 2.4898 & 0.3501 \\
    + cond. unsampling & 19.2002 & 0.0707 & 0.0214 & \underline{0.7786} & \underline{0.1181} & 2.5074 & 0.3499 \\
    + parts composition & \textbf{17.1890} & \textbf{0.0544} & \textbf{0.0156} & 0.7737 & 0.1142 & \underline{2.5092} & \underline{0.3410} \\
    + logo restoration & \underline{18.9950} & \underline{0.0606} & \underline{0.0245} & 0.7708 & 0.1177 & \textbf{2.5096} & \textbf{0.3353} \\
    \midrule
    \rowcolor{white} \multicolumn{8}{c}{\textit{DressCode (upper-body with brand logos)}} \\
    \midrule
    Coarse generation & 90.8324 & 0.0746 & 0.0174 & \textbf{0.8324} & \textbf{0.1153} & 2.5927 & 0.2073 \\
    + cond. unsampling & \underline{74.9592} & 0.0711 & 0.0171 & \underline{0.8212} & \underline{0.1272} & \underline{2.6212} & \underline{0.1730} \\
    + parts composition & 75.6367 & \underline{0.0599} & \underline{0.0146} & 0.8174 & 0.1225 & 2.6195 & 0.1479 \\
    + logo restoration & \textbf{55.9897} & \textbf{0.0555} & \textbf{0.0240} & 0.8037 & 0.1242 & \textbf{2.6256} & \textbf{0.2044} \\
    \midrule
    \rowcolor{white} \multicolumn{8}{c}{\textit{VITON-HD}} \\
    \midrule
    Coarse generation & 30.7279 & 0.0741 & 0.0193 & \textbf{0.8290} & \textbf{0.1019} & \textbf{2.5036} & \textbf{0.3483} \\
    + cond. unsampling & 28.1324 & 0.0750 & 0.0206 & \underline{0.8158} & \underline{0.1152} & \underline{2.4959} & \underline{0.3695} \\
    + parts composition & \textbf{21.4977} & \textbf{0.0551} & \textbf{0.0143} & 0.8109 & 0.1100 & 2.4831 & 0.3604 \\
    + logo restoration & \underline{24.5848} & \underline{0.0645} & \underline{0.0244} & 0.8113 & 0.1135 & 2.4947 & 0.3565 \\
    \midrule
    \rowcolor{white} \multicolumn{8}{c}{\textit{VITON-HD (garments with brand logos)}} \\
    \midrule
    Coarse generation & 185.3339 & 0.0717 & 0.0081 & \textbf{0.8056} & \textbf{0.1270} & \underline{2.6458} & \underline{0.0329} \\
    + cond. unsampling & 176.1029 & 0.0762 & 0.0136 & 0.7946 & 0.1405 & 2.6280 & 0.1346 \\
    + parts composition & \underline{166.8071} & \underline{0.0625} & \underline{0.0078} & \underline{0.7968} & \underline{0.1347} & 2.6385 & 0.0366 \\
    + logo restoration & \textbf{99.7761} & \textbf{0.0544} & \textbf{0.0212} & 0.7760 & 0.1352 & \textbf{2.6852} & \textbf{0.0482} \\
    \midrule
    \rowcolor{white} \multicolumn{8}{c}{\textit{OVS (long sleeve garments)}} \\
    \midrule
    Coarse generation & 51.6678 & 0.0694 & 0.0179 & \textbf{0.7598} & \textbf{0.0939} & 2.5076 & 0.2642 \\
    + cond. unsampling & 51.2788 & 0.0667 & 0.0199 & 0.7388 & 0.1093 & 2.4914 & 0.3022 \\
    + parts composition & \underline{45.0715} & \textbf{0.0508} & \textbf{0.0147} & \underline{0.7408} & \underline{0.1059} & 2.5080 & 0.2984 \\
    + logo restoration & \textbf{44.0097} & \underline{0.0564} & \underline{0.0248} & 0.7390 & 0.1094 & \textbf{2.5163} & \textbf{0.2975} \\
    \bottomrule
  \end{tabular}
  \caption{\textbf{Quantitative Ablation Study.} Quantitative results of the pipeline stages on DressCode \citep{DBLP:journals/tog/HeYZYLX24} and VITON-HD \citep{choi2021viton} with the corresponding subsets containing only garments with brand logos. An additional analysis was conducted on the production dataset of OVS. Best results are reported in \textbf{bold}, while second-best are \underline{underlined}.}
  \label{tab:quantitative_evaluation_ablation}
\end{table}

%% file: qualitative_evaluation_ablation.tex
\begin{figure*}[t]
    \centering
    \begin{subfigure}{0.135\linewidth}
        \includegraphics[width=\linewidth, trim={2cm 2.5cm 2cm 2.5cm},clip]{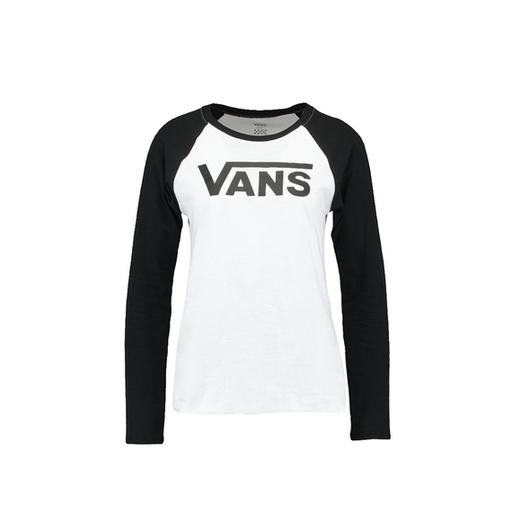}
    \end{subfigure}
    \begin{subfigure}{0.20\linewidth}
        \includegraphics[width=\linewidth, trim={0 3cm 0 3cm},clip]{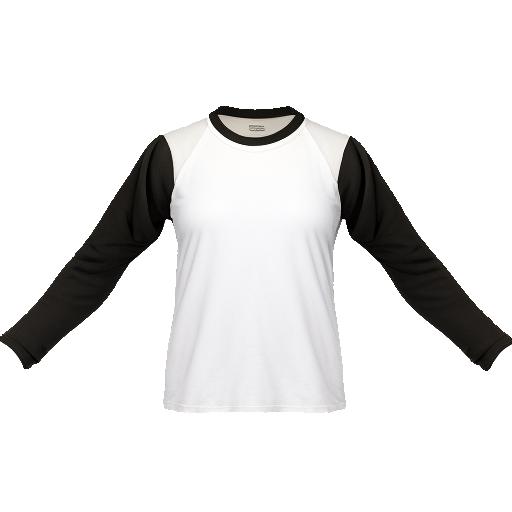}
    \end{subfigure}
    \begin{subfigure}{0.20\linewidth}
        \includegraphics[width=\linewidth, trim={0 3cm 0 3cm},clip]{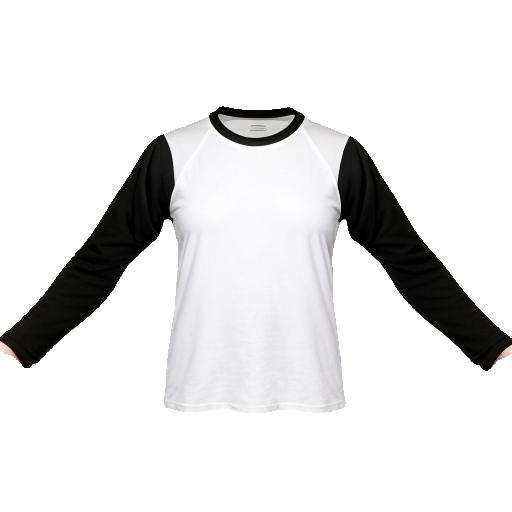}
    \end{subfigure}
    \begin{subfigure}{0.20\linewidth}
        \includegraphics[width=\linewidth, trim={0 3cm 0 3cm},clip]{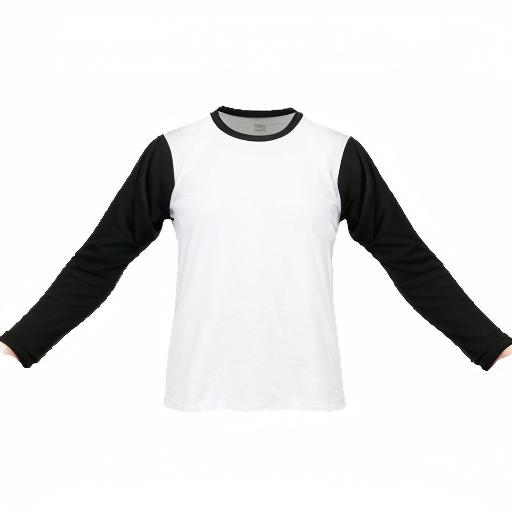}
    \end{subfigure}
    \begin{subfigure}{0.20\linewidth}
        \includegraphics[width=\linewidth, trim={0 6cm 0 6cm},clip]{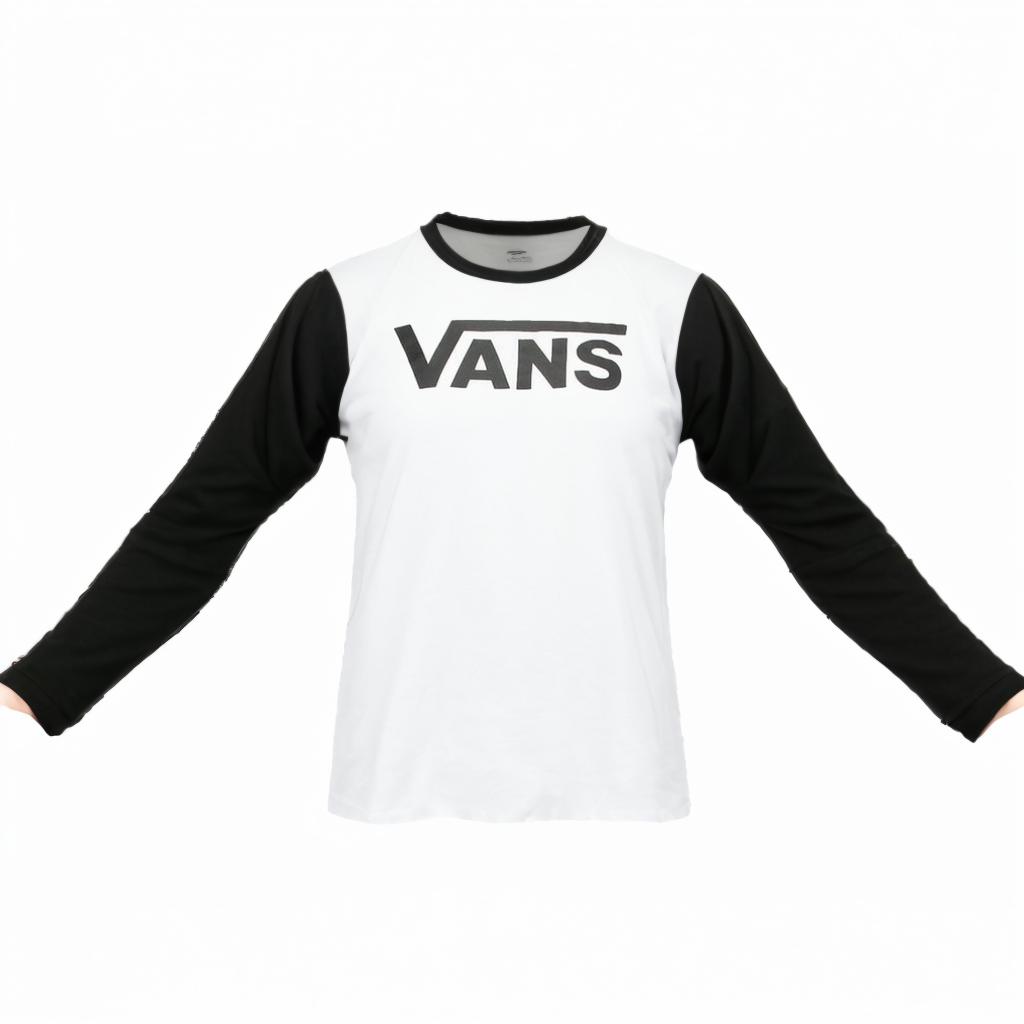}
    \end{subfigure} \\
    \begin{subfigure}{0.135\linewidth}
        \includegraphics[width=\linewidth, trim={2cm 2.5cm 2cm 2.5cm},clip]{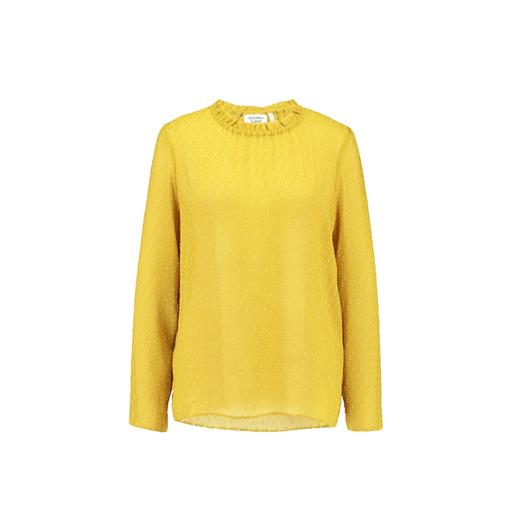}
    \end{subfigure}
    \begin{subfigure}{0.20\linewidth}
        \includegraphics[width=\linewidth, trim={0 3cm 0 3cm},clip]{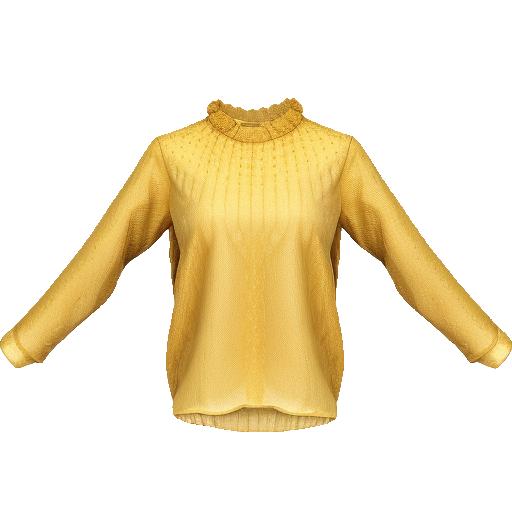}
    \end{subfigure}
    \begin{subfigure}{0.20\linewidth}
        \includegraphics[width=\linewidth, trim={0 3cm 0 3cm},clip]{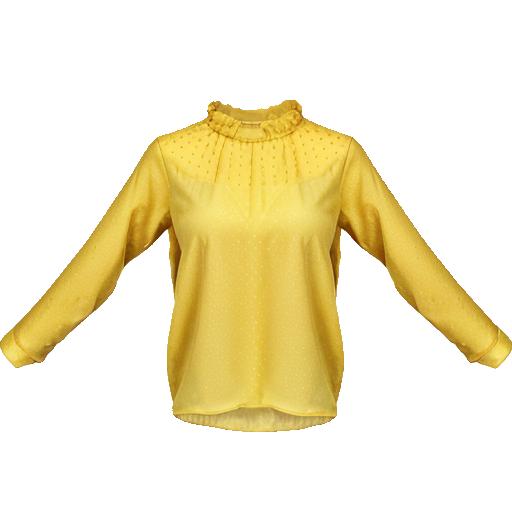}
    \end{subfigure}
    \begin{subfigure}{0.20\linewidth}
        \includegraphics[width=\linewidth, trim={0 3cm 0 3cm},clip]{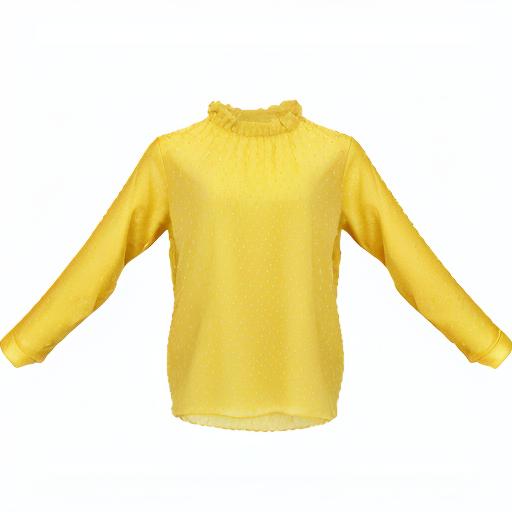}
    \end{subfigure}
    \begin{subfigure}{0.20\linewidth}
        \includegraphics[width=\linewidth, trim={0 6cm 0 6cm},clip]{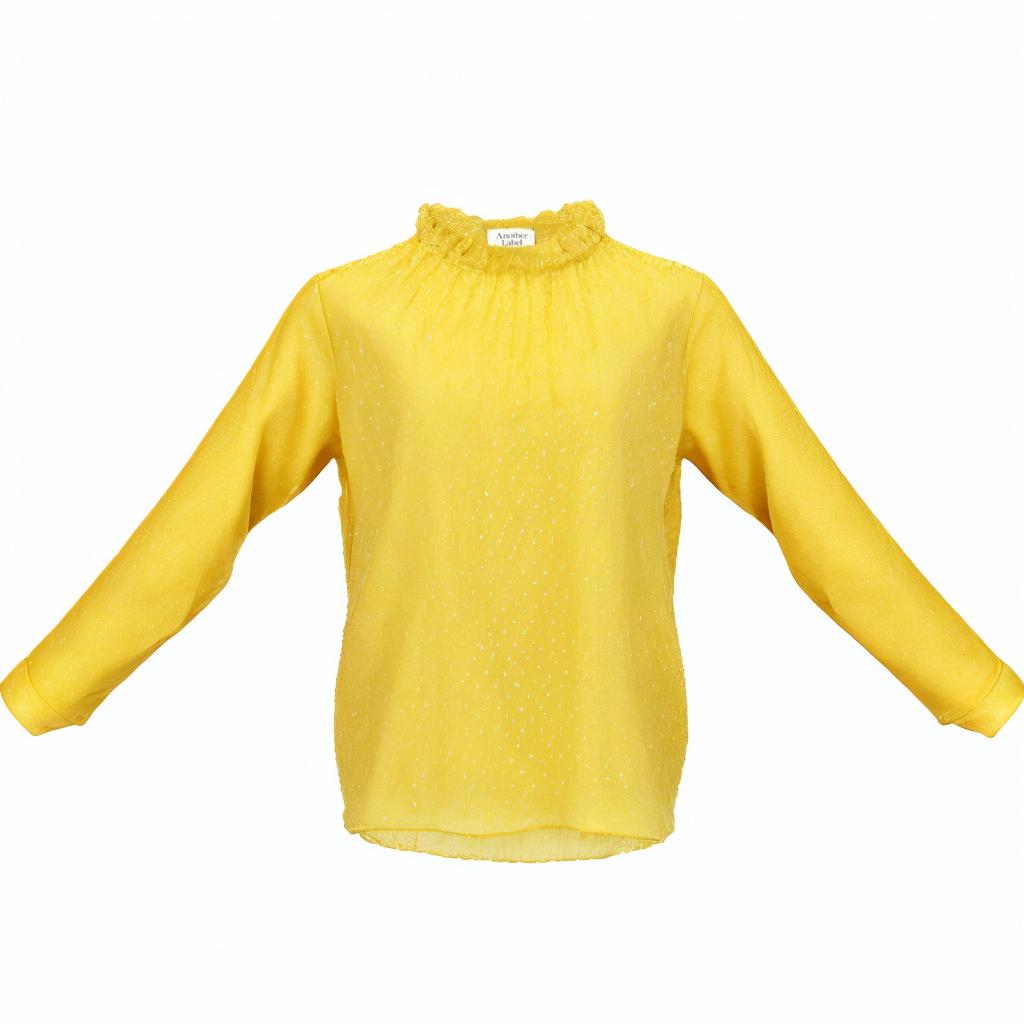}
    \end{subfigure} \\
    \begin{subfigure}{0.135\linewidth}
        \includegraphics[width=\linewidth, trim={2cm 2.5cm 2cm 2.5cm},clip]{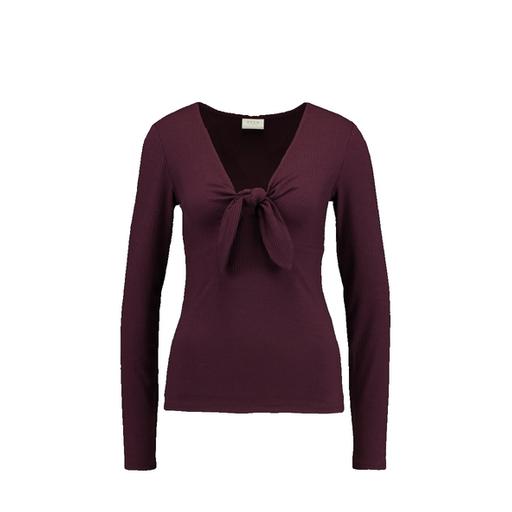}
        \caption{Reference}
    \end{subfigure}
    \begin{subfigure}{0.20\linewidth}
        \includegraphics[width=\linewidth, trim={0 3cm 0 3cm},clip]{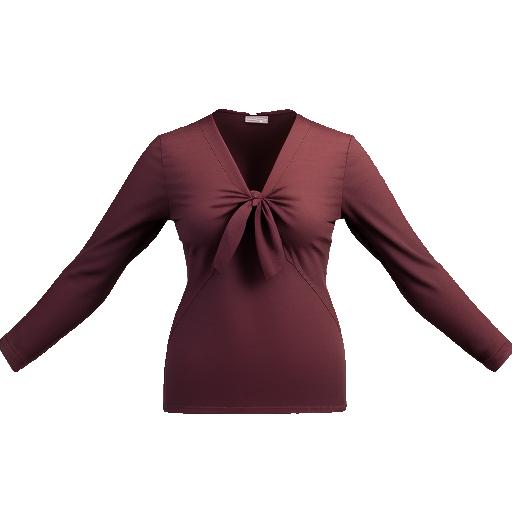}
        \caption{+ Coarse gen.}
    \end{subfigure}
    \begin{subfigure}{0.20\linewidth}
        \includegraphics[width=\linewidth, trim={0 3cm 0 3cm},clip]{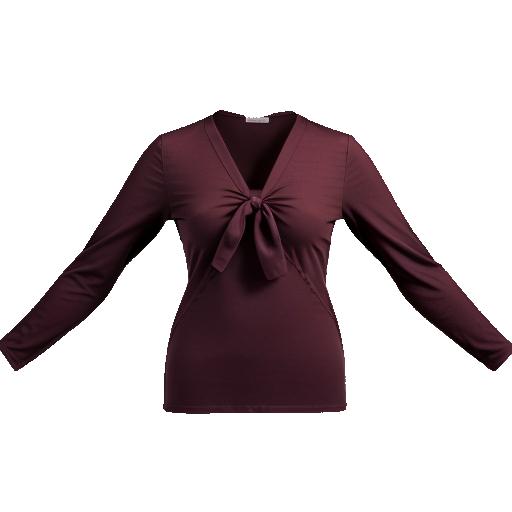}
        \caption{+ Cond. unsamp.}
    \end{subfigure}
    \begin{subfigure}{0.20\linewidth}
        \includegraphics[width=\linewidth, trim={0 3cm 0 3cm},clip]{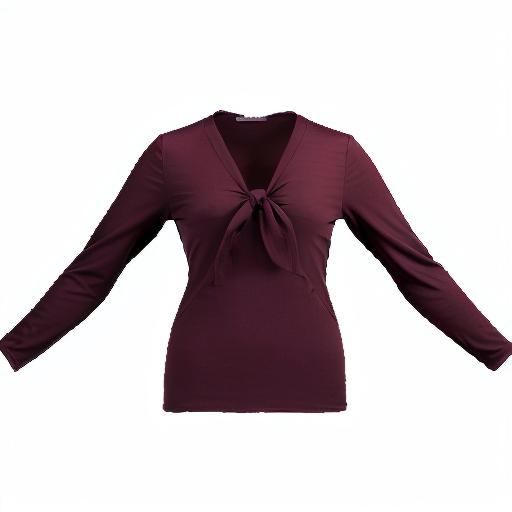}
        \caption{+ Parts compos.}
    \end{subfigure}
    \begin{subfigure}{0.20\linewidth}
        \includegraphics[width=\linewidth, trim={0 6cm 0 6cm},clip]{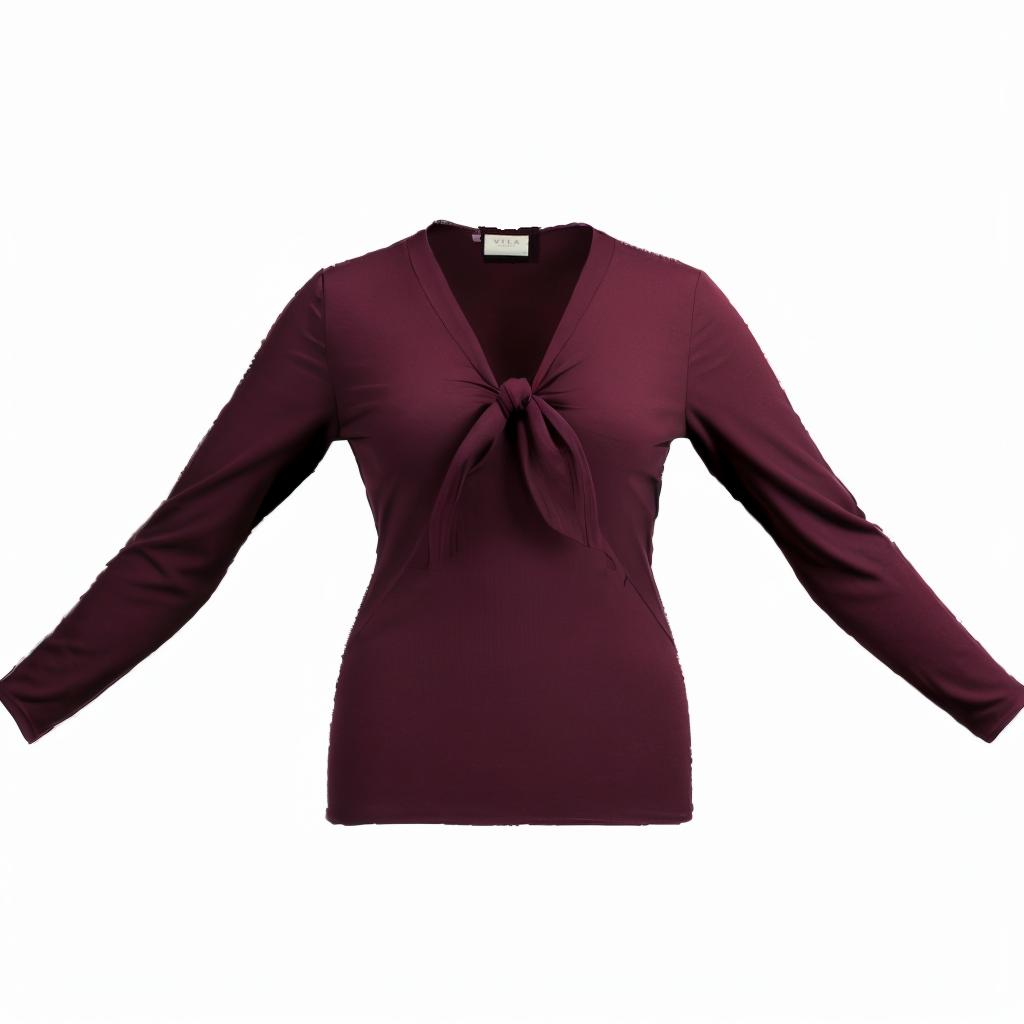}
        \caption{+ Logo restor.}
    \end{subfigure}
    \caption{\textbf{Qualitative Ablation Study.} The figure illustrates the qualitative ablation study considering incremental stages of the pipeline. The clothing images are sampled from VITON-HD~\citep{choi2021viton} dataset.}
    \label{fig:qualitative_evaluation_ablation}
\end{figure*}

%% file: pareto_sota.tex
\newcommand{\datasetDresscode}{pareto_sota_DressCode-long-sleeves-w-logo.csv}
\newcommand{\datasetVitonHD}{pareto_sota_VITON-HD-long-sleeves-w-logo.csv}
\newcommand{\markerSize}{1.2}
\definecolor{colorMasaCtrl}{HTML}{000000}
\definecolor{colorNullTextInversion}{HTML}{EE7733}
\definecolor{colorTIC}{HTML}{009988}
\definecolor{colorFPE}{HTML}{88CCEE}
\definecolor{colorControlNet}{HTML}{882255}
\definecolor{colorOurs}{HTML}{DDCC77}

\begin{figure}
    \begin{tikzpicture}
        \begin{axis}[
            xlabel={1 - FID\textsubscript{normalized}},
            ylabel={SSIM},
            grid=both,
            width=\linewidth,
            height=5.5cm,
            legend cell align=left,
            legend pos=south east,
            legend style={font=\scriptsize},
            tick label style={font=\scriptsize},
            label style={font=\footnotesize},
            xmin=-0.04,
            xmax=0.95,
            ymin=0.14,
            ymax=0.85
        ]

        \addplot[
            scatter/classes={
                MasaCtrl={mark=triangle*,draw=black,fill=colorMasaCtrl,scale=\markerSize},
                NullTextInversion={mark=square*,draw=black,fill=colorNullTextInversion,scale=\markerSize},
                TIC={mark=diamond*,draw=black,fill=colorTIC,scale=\markerSize},
                FPE={mark=*,draw=black,fill=colorFPE,scale=\markerSize},
                ControlNet={mark=otimes*,draw=black,fill=colorControlNet,scale=\markerSize},
                ours={mark=pentagon*,draw=black,fill=colorOurs,scale=\markerSize}
            },
            scatter,only marks,
            scatter src=explicit symbolic,
            error bars/.cd,
            x dir=both, x explicit,
            y dir=both, y explicit
        ] 
        table[
            x = fid,
            y = ssim_mean,
            col sep = comma,
            meta = method
        ] {\datasetDresscode};

        \addplot[
            blue, thick
        ]
        table[
            x = fid,
            y = ssim_mean,
            col sep = comma
        ] {\datasetDresscode};


        \addplot[
            red, thick
        ]
        table[
            x = fid,
            y = ssim_mean,
            col sep = comma
        ] {\datasetVitonHD};

        \addplot[
            scatter/classes={
                MasaCtrl={mark=triangle*,draw=colorMasaCtrl,fill=white,line width=1.25pt,scale=\markerSize},
                NullTextInversion={mark=square*,draw=colorNullTextInversion,fill=white,line width=1.25pt,scale=\markerSize},
                TIC={mark=diamond*,draw=colorTIC,fill=white,line width=1.25pt,scale=\markerSize},
                FPE={mark=*,draw=colorFPE,fill=white,line width=1.25pt,scale=\markerSize},
                ControlNet={mark=otimes*,draw=colorControlNet,fill=white,line width=1.25pt,scale=\markerSize},
                ours={mark=pentagon*,draw=colorOurs,fill=white,line width=1.25pt,scale=\markerSize}
            },
            scatter,only marks,
            scatter src=explicit symbolic,
            error bars/.cd,
            x dir=both, x explicit,
            y dir=both, y explicit
        ] 
        table[
            x = fid,
            y = ssim_mean,
            col sep = comma,
            meta = method
        ] {\datasetVitonHD};

        \legend{
            MasaCtrl,
            Null-Text Inv.,
            TIC,
            FPE,
            ControlNet,
            Ours,
            DressCode,
            VITON-HD
        }
        
        \end{axis}
    \end{tikzpicture}
    \caption{\textbf{State-of-the-art Pareto fronts} for DressCode with logos only (represented with filled markers), and VITON-HD with logos only (represented with empty markers). Best models are located in the top right corner.}
    \label{fig:pareto_quantitative_evaluation}
\end{figure}

%% file: runtime_analysis.tex
\begin{table}[t]
    \centering
    \scriptsize
    \setlength{\tabcolsep}{3pt}
    \rowcolors{2}{}{gray!20}
    \begin{tabular}{@{}lccccc@{}}
      \toprule
      \textbf{Method} & \multicolumn{2}{c}{\textbf{Execution time} (seconds $\downarrow$)} & \multicolumn{3}{c}{\textbf{VRAM usage} (Megabytes $\downarrow$)} \\
      \cmidrule(r){2-3} \cmidrule(r){4-6}
      \rowcolor{white} & \textbf{Mean} & \textbf{Std} & \textbf{Mean} & \textbf{Std} & \textbf{Max} \\
      \midrule
      \rowcolor{white} & \multicolumn{5}{c}{\textit{OVS (long sleeve garments)}} \\
      \midrule
      MasaCtrl \citep{DBLP:conf/iccv/CaoWQSQZ23} & 16.1038 & 1.8893 & 13596.0182 & 750.5771 & 14808.6250 \\
      Null-text Inv. \citep{DBLP:conf/cvpr/MokadyHAPC23} & 27.6974 & 8.5677 & 12262.2553 & 586.8886 & \underline{12838.6250} \\
      TIC \citep{DBLP:conf/aaai/DuanCK0FFH24} & \underline{10.1216} & \underline{1.2170} & 20158.4967 & 510.2891 & 22566.6250 \\
      FPE \citep{DBLP:conf/cvpr/Liu0CJH24} & 18.6562 & 2.5453 & 15619.8162 & 1346.4427 & 20280.6250 \\
      ControlNet \citep{DBLP:conf/iccv/ZhangRA23} & \textbf{2.2278} & \textbf{0.2354} & \textbf{3785.1770} & \textbf{271.6237} & \textbf{3959.3125} \\
      Ours & 22.9994 & 3.3531 & \underline{11964.4932} & \underline{576.0706} & 15137.3125 \\
      \bottomrule
    \end{tabular}
    \caption{\textbf{Quantitative Runtime Analysis.} Runtime analysis conducted on a real-world production dataset from OVS, consisting of 861 long-sleeve product images collected over the past year. Execution time is expressed in seconds and VRAM usage in megabytes (MB). Best results are reported in \textbf{bold}, while second-best are \underline{underlined}.}
    \label{tab:runtime_analysis}
\end{table}

%% file: vram_usage_ours.tex
\newcommand{\vramUsageData}{vram_usage_ours_table.csv}
\definecolor{customColor}{HTML}{1F77B4}

\begin{figure}
    \begin{tikzpicture}
        \begin{axis}[
            xlabel={Execution time (seconds)},
            ylabel={VRAM usage (MB)},
            grid=both,
            width=\linewidth,
            height=5.5cm,
            tick label style={font=\scriptsize},
            label style={font=\footnotesize},
            xmin=-1.0,
            xmax=27.5
        ]

        \addplot[name path=lower, fill=none, draw=none]
        table[
            x = time,
            y = ci_lower,
            col sep = comma
        ] {\vramUsageData};

        \addplot[name path=upper, fill=none, draw=none]
        table[
            x = time,
            y = ci_upper,
            col sep = comma
        ] {\vramUsageData};

        \addplot[customColor, opacity=0.2] fill between[of=lower and upper];

        \addplot [customColor, line width = 1.5px]
        table [
            x = time,
            y = mean,
            col sep = comma
        ]{\vramUsageData};
        
        \end{axis}
    \end{tikzpicture}
    \caption{{\textbf{VRAM usage over time.} The figure illustrates the VRAM consumption profile over time, during inference on a real-world production dataset from OVS, comprising 861 long-sleeve product images collected over the past year. The plotted line denotes the mean VRAM usage, while the shaded region represents the 95\% confidence interval.}}
    \label{fig:runtime_analysis_vram_usage}
\end{figure}

%% file: qualitative_evaluation_fashiontryon.tex
\begin{figure*}[t]
  \centering
  \resizebox{\textwidth}{!}{
      \includegraphics[height=2cm, width=1.65cm]{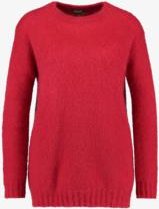}
      \includegraphics[height=2cm]{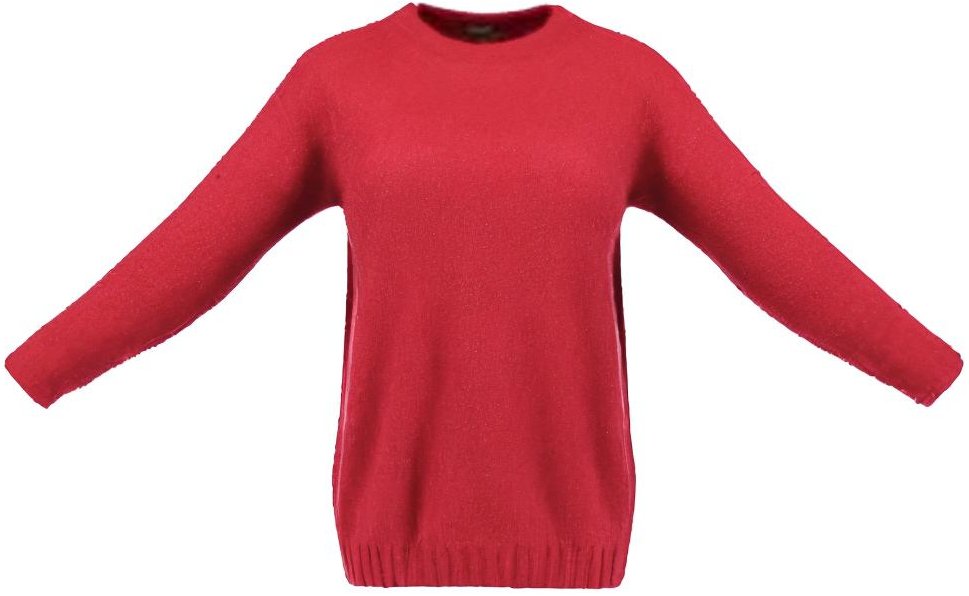}
      \includegraphics[height=2cm, width=1.65cm]{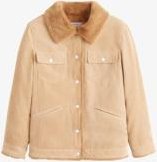}
      \includegraphics[height=2cm]{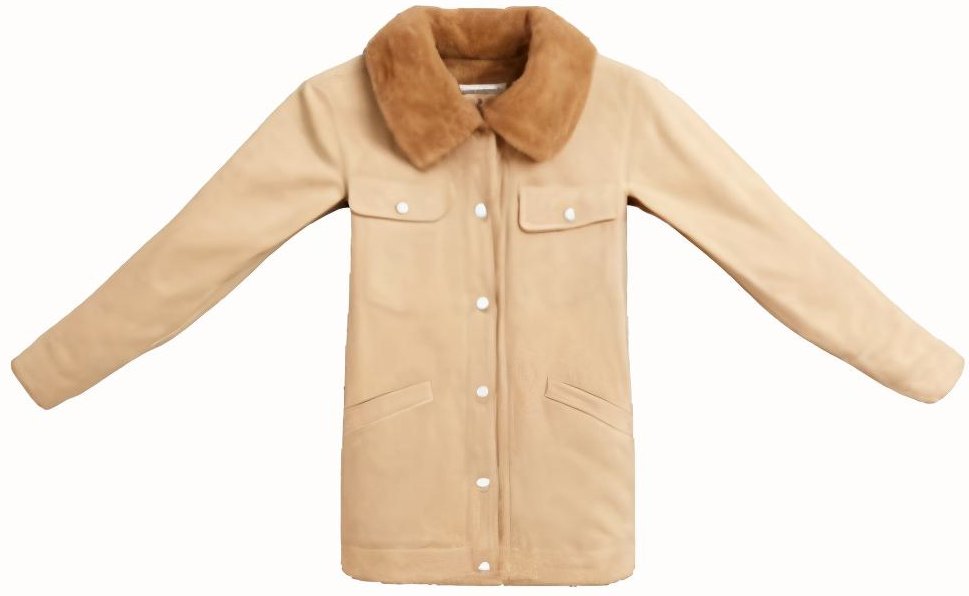}
      \includegraphics[height=2cm, width=1.65cm]{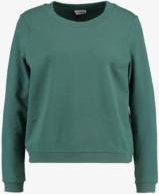}
      \includegraphics[height=2cm]{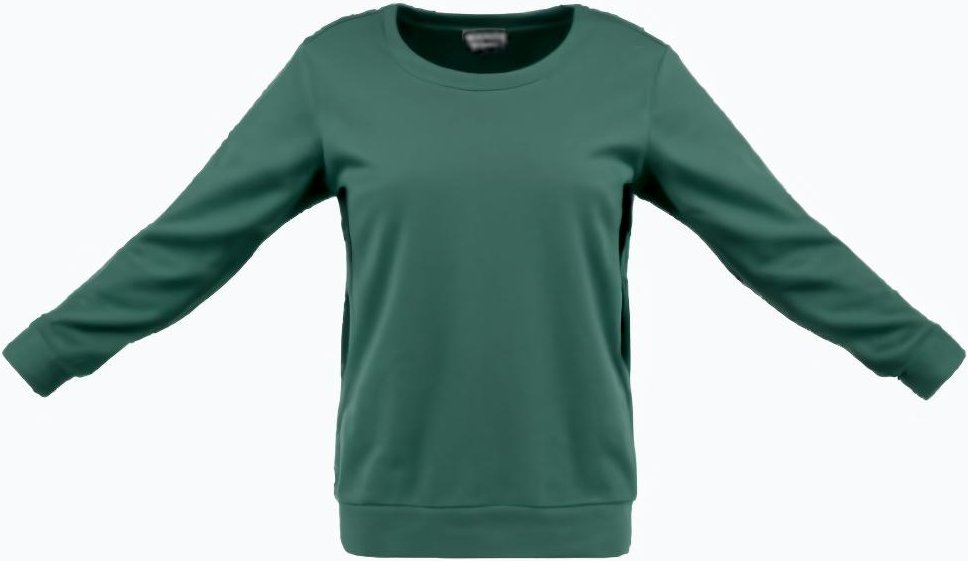}
  } \\
  \resizebox{\textwidth}{!}{
      \includegraphics[height=2cm, width=1.65cm]{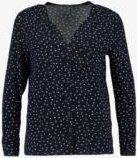}
      \includegraphics[height=2cm]{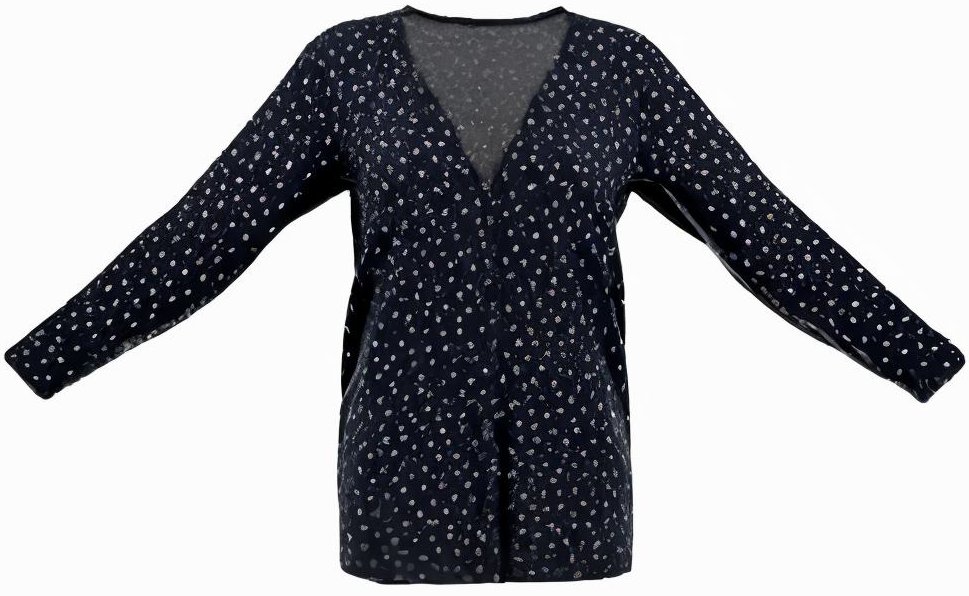}
      \includegraphics[height=2cm, width=1.65cm]{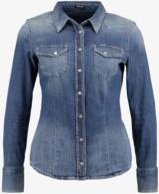}
      \includegraphics[height=2cm]{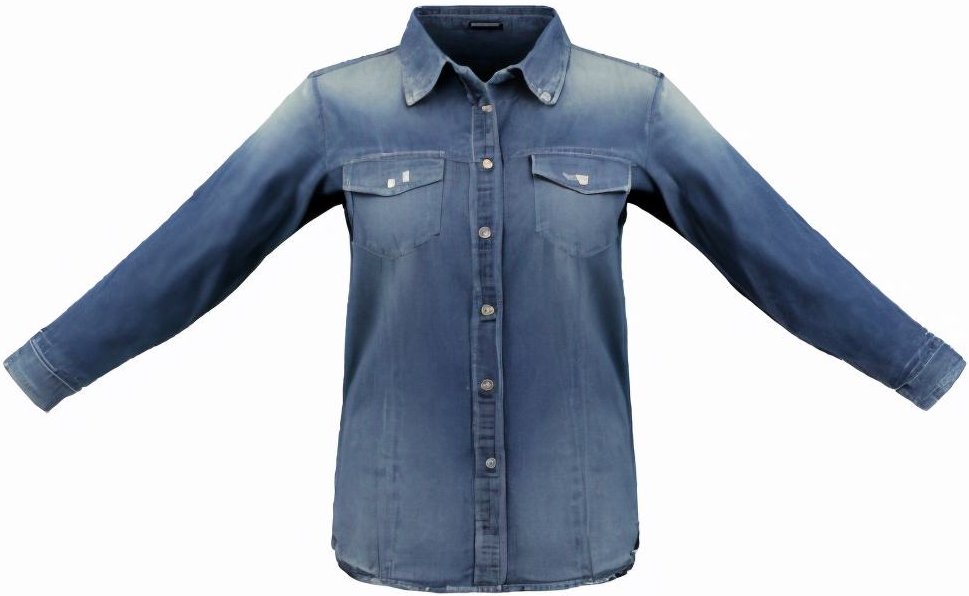}
      \includegraphics[height=2cm, width=1.65cm]{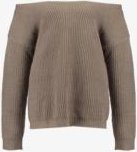}
      \includegraphics[height=2cm]{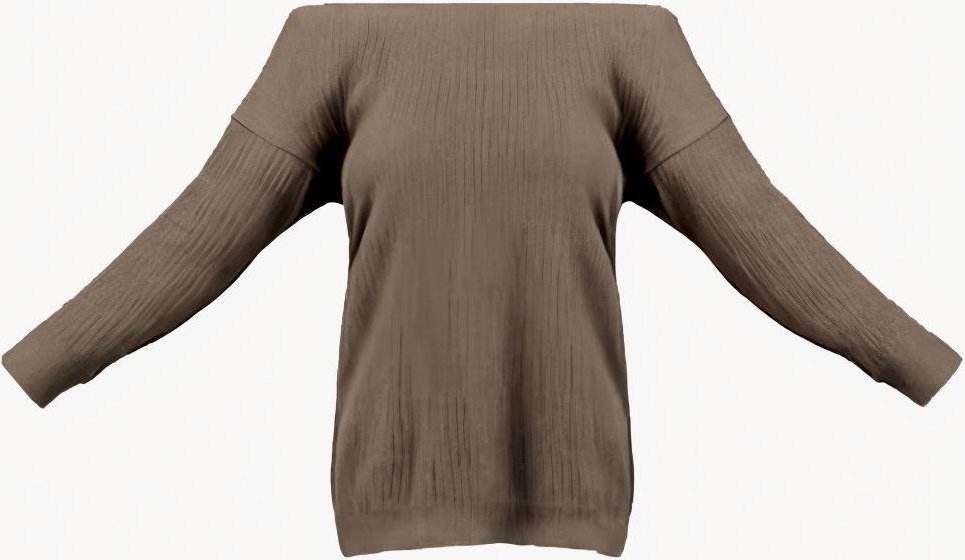}
  } \\
  \resizebox{\textwidth}{!}{
      \includegraphics[height=2cm, width=1.65cm]{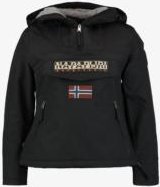}
      \includegraphics[height=2cm]{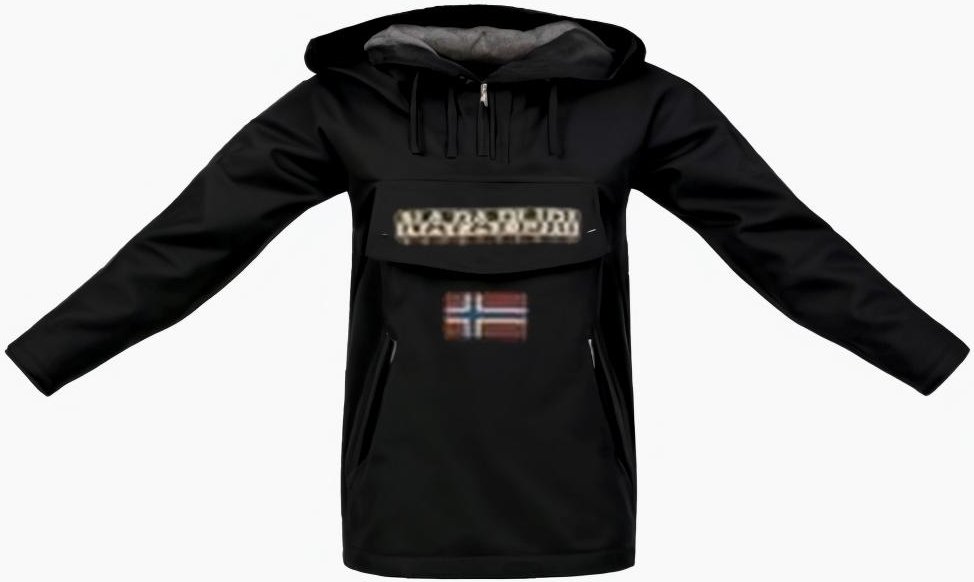}
      \includegraphics[height=2cm, width=1.65cm]{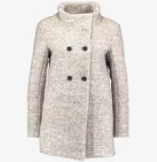}
      \includegraphics[height=2cm]{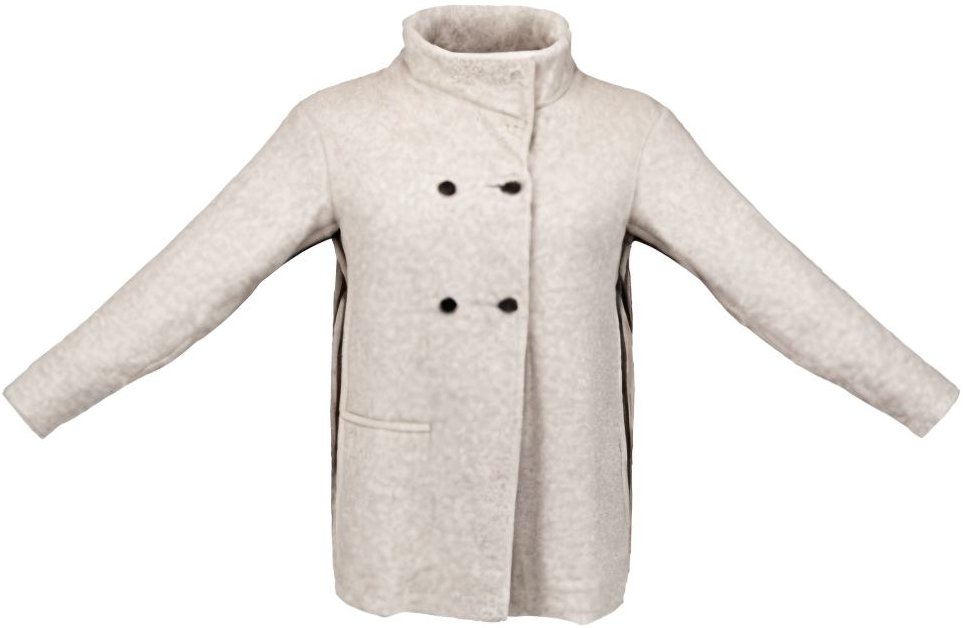}
      \includegraphics[height=2cm, width=1.65cm]{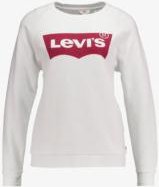}
      \includegraphics[height=2cm]{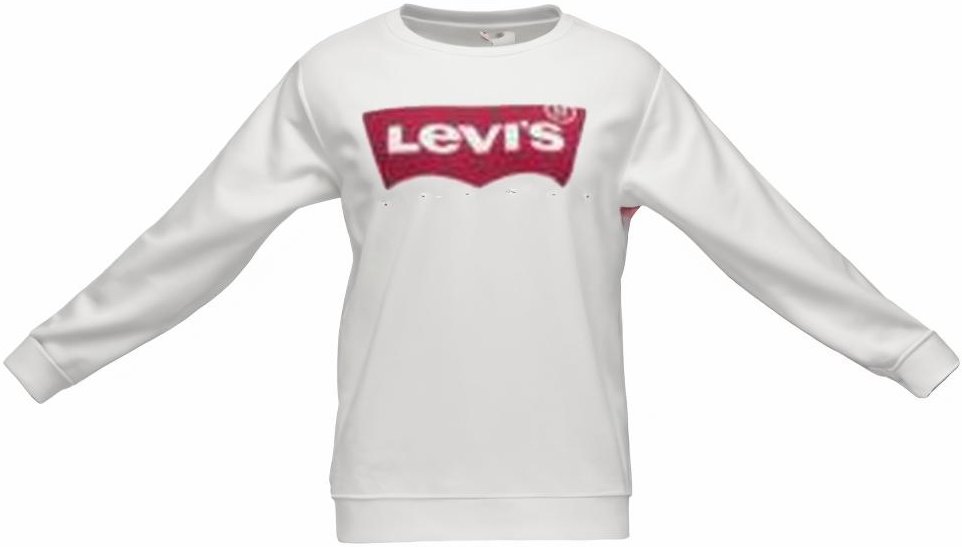}
  } \\
  \caption{\review{\textbf{Qualitative Evaluation on FashionTryOn.} Visual results obtained using FashionRepose on samples from the FashionTryOn dataset~\citep{DBLP:conf/mm/ZhengSCHCN19}. Each pair shows the original still-life garment (left) and the reposed output (right).}}
  \label{fig:qualitative_evaluation_fashiontryon}
\end{figure*}

%% file: pareto_ablation.tex
\renewcommand{\datasetDresscode}{pareto_ablation_DressCode-long-sleeves-w-logo.csv}
\renewcommand{\datasetVitonHD}{pareto_ablation_VITON-HD-long-sleeves-w-logo.csv}
\renewcommand{\markerSize}{1.2}
\definecolor{colorCoarseGeneration}{HTML}{000000}
\definecolor{colorConditionedUnsampling}{HTML}{88CCEE}
\definecolor{colorPartsComposition}{HTML}{009988}
\definecolor{colorLogoRestoration}{HTML}{DDCC77}

\begin{figure}
    \begin{tikzpicture}
        \begin{axis}[
            xlabel={1 - FID\textsubscript{normalized}},
            ylabel={SSIM},
            grid=both,
            width=\linewidth,
            height=5.5cm,
            legend cell align=left,
            legend pos=south east,
            legend style={font=\scriptsize},
            tick label style={font=\scriptsize},
            label style={font=\footnotesize},
            xmin=-0.02,
            xmax=0.5,
            ymin=0.3,
            ymax=0.9
        ]

        \addplot[
            scatter/classes={
                ipadapter={mark=triangle*,draw=black,fill=colorCoarseGeneration,scale=\markerSize},
                conditioned_unsampling={mark=square*,draw=black,fill=colorConditionedUnsampling,scale=\markerSize},
                parts_composition={mark=diamond*,draw=black,fill=colorPartsComposition,scale=\markerSize},
                logo_restoration={mark=pentagon*,draw=black,fill=colorLogoRestoration,scale=\markerSize}
            },
            scatter,only marks,
            scatter src=explicit symbolic
        ] 
        table[
            x = fid,
            y = SSIM mean,
            col sep = comma,
            meta = Pipeline stage
        ] {\datasetDresscode};

        \addplot[
            blue, thick
        ]
        table[
            x = fid,
            y = SSIM mean,
            col sep = comma
        ] {\datasetDresscode};


        \addplot[
            red, thick
        ]
        table[
            x = fid,
            y = SSIM mean,
            col sep = comma
        ] {\datasetVitonHD};

        \addplot[
            scatter/classes={
                ipadapter={mark=triangle*,draw=colorCoarseGeneration,fill=white,line width=1.25pt,scale=\markerSize},
                conditioned_unsampling={mark=square*,draw=colorConditionedUnsampling,fill=white,line width=1.25pt,scale=\markerSize},
                parts_composition={mark=diamond*,draw=colorPartsComposition,fill=white,line width=1.25pt,scale=\markerSize},
                logo_restoration={mark=pentagon*,draw=colorLogoRestoration,fill=white,line width=1.25pt,scale=\markerSize}
            },
            scatter,only marks,
            scatter src=explicit symbolic
        ] 
        table[
            x = fid,
            y = SSIM mean,
            col sep = comma,
            meta = Pipeline stage
        ] {\datasetVitonHD};


        \addplot [
            domain=-1:1,
            samples=2,
            draw=none,
            fill=gray,
            fill opacity=0.3,
            pattern=north east lines,
            pattern color=black
        ] {0.6} |- (0,0) -- cycle;

        \addplot [
            domain=-1:1,
            samples=2,
            color=red,
            dashed,
            ultra thick
        ] {0.6};

        \legend{
            Coarse Generation,
            Conditioned Unsampling,
            Parts Composition,
            Logo Restoration,
            DressCode,
            VITON-HD
        }
        
        \end{axis}
    \end{tikzpicture}
    \caption{\textbf{Ablation Pareto fronts} for DressCode with logos only (represented with filled markers), and VITON-HD with logos only (represented with empty markers). We consider the repose task successful when SSIM score is $>$ 0.6.}
    \label{fig:pareto_ablation_evaluation}
\end{figure}

%% file: 4_limitations.tex
\section{Limitations and Future Work}
\label{sec:conclusion}


While \textsc{FashionRepose} demonstrates strong performance in both identity preservation and pose alignment, certain limitations persist (see Fig.~\ref{fig:limitations}). Most notably, the pipeline struggles with garments featuring highly complex patterns or textures. In such cases, the coarse generation and conditioned unsampling stages may introduce distortions or inconsistencies across garment regions, as illustrated in the left panel of Figure~\ref{fig:limitations}, where duplicated sleeves with mismatched textures emerge. Additionally, the accuracy of the garment parts-composition step is contingent on reliable mask computation. Imprecise mask boundaries, especially when occlusions or visual ambiguities are present, can lead to visible seams, unblended transitions, or small gaps in the composed image. Another notable shortcoming arises in the logo restoration module. As shown in the right panel of Figure~\ref{fig:limitations}, semantic misclassification can occur when other elements are erroneously identified as logos, resulting in misplaced reinjections that compromise visual realism.

These limitations highlight the need for improved robustness in logo detection and enhanced generalization to garments with intricate visual properties. Future work will address these challenges through the refinement of logo classification strategies, improved segmentation fidelity, and broader garment-type support.

\input{limitations}

%% file: limitations.tex
\begin{figure}[t]
  \centering
  \resizebox{1.0\textwidth}{!}{
      \includegraphics[height=2cm, trim={5cm 2.5cm 5cm 3.8cm}, clip]{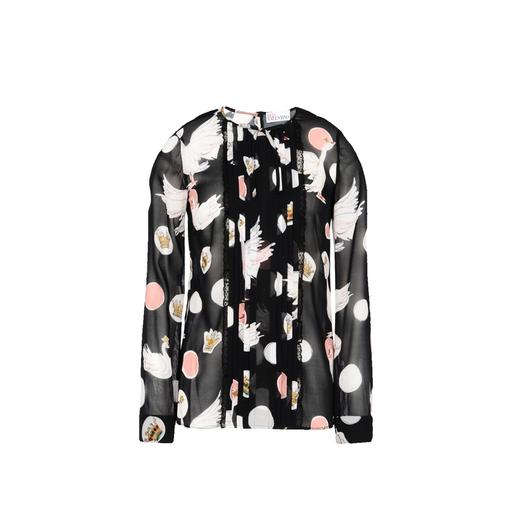}
      \includegraphics[height=2cm, trim={0cm 3.2cm 0cm 7.8cm}, clip]{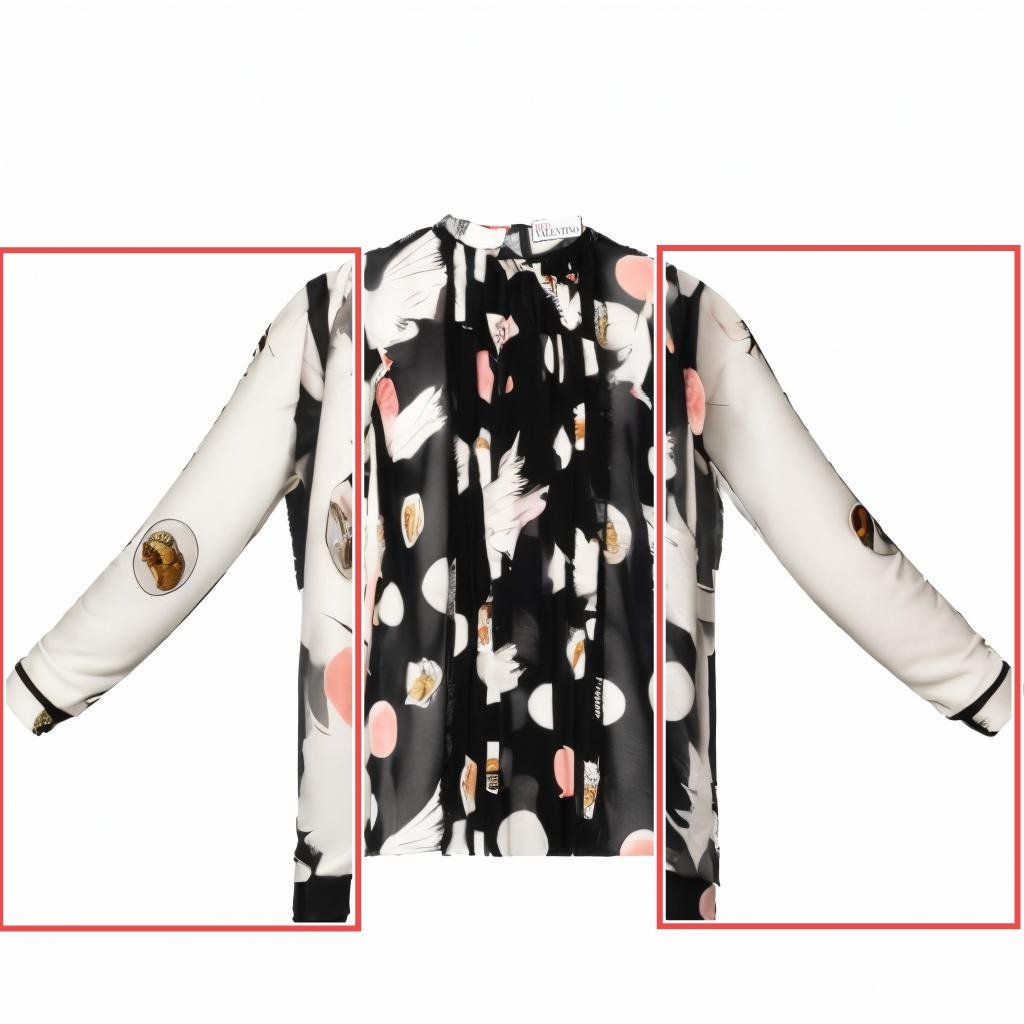}
      \includegraphics[height=2cm, trim={5cm 3.5cm 5cm 2.5cm}, clip]{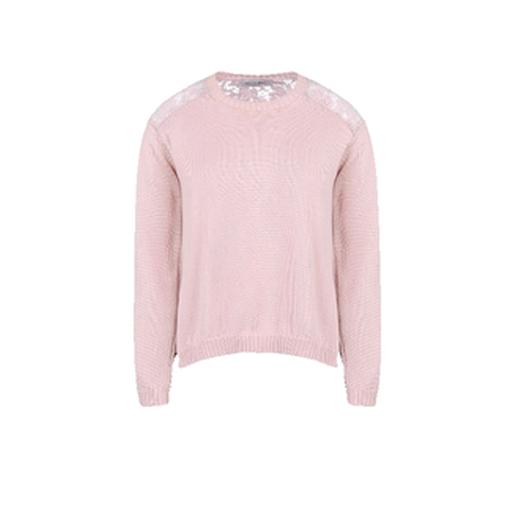}
      \includegraphics[height=2cm, trim={0cm 10cm 0cm 2.5cm}, clip]{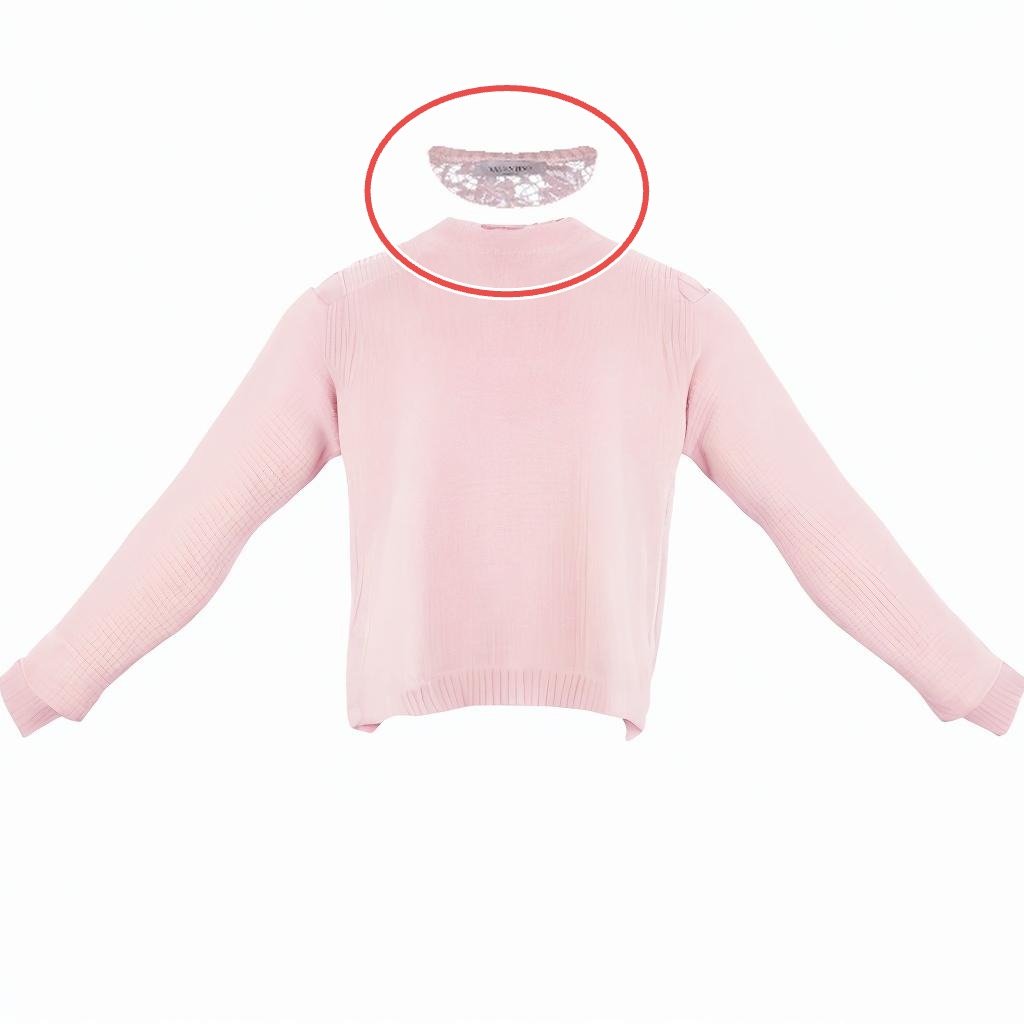}
  }
  \caption{\textbf{Limitations.} Red boxes highlight visual artifacts introduced by the pipeline. On the left panel, duplicated sleeves with mismatched textures are generated due to the garment's complex shape and pattern. On the right panel, the neck region is mistakenly identified as a logo, resulting in incorrect placement during the logo injection phase.}
  \label{fig:limitations}
\end{figure}

%% file: 5_conclusion.tex
\section{Conclusion}
\label{sec:conclusion}

This work presents \textsc{FashionRepose}, a novel training-free and zero-shot pipeline for garment pose alignment, specifically designed for long-sleeved upper-body garments in still-life fashion imagery. The proposed method enables non-rigid pose transformations while preserving key visual attributes such as shape, texture, and branding. By combining pretrained diffusion models with computer vision techniques in a modular architecture, the system eliminates the need for model fine-tuning or specialized training data, ensuring rapid deployment and broad applicability.

Extensive evaluations across public datasets and real-world production data confirm the effectiveness of the approach. \textsc{FashionRepose} demonstrates strong performance on perceptual and structural metrics, particularly in scenarios involving branded garments, and consistently preserves garment identity across pose variations. An ablation study highlights the critical role of each pipeline component, particularly the composition and logo-restoration modules, in achieving high-quality and coherent results.

While a few edge cases involving complex textures or segmentation inaccuracies remain, they do not significantly affect the pipeline’s practical utility. Future refinements may further enhance robustness and generalization across a wider variety of garment types.


